\pdfoutput=1
\documentclass[11pt]{article}
\usepackage[preprint]{acl}
\usepackage{times}
\usepackage{latexsym}
\usepackage[T1]{fontenc}
\usepackage[utf8]{inputenc}
\usepackage{microtype}
\usepackage{booktabs}
\usepackage{graphicx}
\usepackage{xcolor}
\definecolor{PlotBlue}{HTML}{0072B2}
\definecolor{RepairGreen}{HTML}{009E73}
\definecolor{HarmOrange}{HTML}{D55E00}
\definecolor{PlotAmber}{HTML}{B07500}
\usepackage{amsmath}
\usepackage{array}
\usepackage{tabularx}
\usepackage{tikz}
\renewcommand{\dbltopfraction}{0.88}
\renewcommand{\dblfloatpagefraction}{0.60}

\makeatletter
\newenvironment{promptblock}%
  {\@beginparpenalty\@M
   \list{}{\setlength{\leftmargin}{0.8em}\setlength{\rightmargin}{0pt}%
     \setlength{\topsep}{3pt}\setlength{\partopsep}{0pt}%
     \setlength{\parsep}{0pt}\setlength{\itemsep}{0pt}}%
   \item\relax\fontsize{8}{9.5}\selectfont\ttfamily\raggedright
   \microtypesetup{protrusion=false}%
   \setlength{\parskip}{0pt}\setlength{\parindent}{0pt}\interlinepenalty\@M}%
  {\endlist}

\newcommand{\pline}[1]{{\parindent=0pt\parshape=2
  \@totalleftmargin \linewidth
  \dimexpr\@totalleftmargin+1.2em\relax \dimexpr\linewidth-1.2em\relax
  #1\par}\nobreak}
\makeatother

\title{Return or Revise?\\
Learning When Revision Helps Retrieval-Augmented QA}
\author{Nicholas Kashani Motlagh \\
  DCS Corp \\
  {\small\texttt{nicholas.kashani\_motlagh.ctr@us.af.mil}}
  \And Tim Anderson \\
  Air Force Research Laboratory \\
  {\small\texttt{timothy.anderson.20@us.af.mil}}
  \AND Jeremy Gwinnup \\
  Air Force Research Laboratory \\
  {\small\texttt{jeremy.gwinnup.1@us.af.mil}}
  \And Grant Erdmann \\
  Air Force Research Laboratory \\
  {\small\texttt{grant.erdmann@us.af.mil}}}
\hypersetup{
  pdftitle={Return or Revise? Learning When Revision Helps Retrieval-Augmented QA},
  pdfauthor={Nicholas Kashani Motlagh, Tim Anderson, Jeremy Gwinnup, Grant Erdmann}
}

\begin{document}
\maketitle
\begingroup
\NoHyper
\renewcommand{\thefootnote}{}%
\footnotetext{This work is sponsored by the Air Force Research Laboratory under
Air Force contract FA2384-26-F-B001. Opinions, interpretations, conclusions and
recommendations are those of the authors and are not necessarily endorsed by
the United States Government. Cleared for public release on 27 Aug 2026. Case
number AFRL-2026-1477.}%
\endNoHyper
\endgroup

\begin{abstract}
We consider the decision of whether to return an existing draft answer or
revise it using retrieved evidence, as in answer-revision systems. Draft
confidence estimates whether the current answer is correct, but the decision
requires estimating the effect of a specified revision. For offline training
and evaluation, we grade both the returned draft and its candidate revision
under the same correctness judge, which makes repair, harm, and the gap to an
oracle observable. We call this paired effect its \emph{recoverability}, and we
train policies to predict it before revision. On 25,870 held-out open-domain
questions across three revision setups, a scorer trained on the paired outcome
has greater area under the accuracy--revision-rate curve than a matched
draft-correctness scorer in all nine Llama setup--seed fits, and gains
0.23--0.68 accuracy points on average at development-selected thresholds, a
difference significant across training runs only for dense retrieval. The
resulting policy improves on always revising and on average closes more than a
third of the oracle gap, although it still applies 38--46\% of the harmful
revisions. When a draft-free standard-RAG answer is also available, however,
choosing between the draft and that answer is stronger by about two points for
Llama and four for OLMo, and adding candidate revision as a third option yields
no significant gain. Recoverability describes one revision; its value as an
available action also depends on the alternatives.
\end{abstract}

\section{Introduction}

Standard retrieval-augmented generation (RAG) answers a question from retrieved
evidence~\citep{lewis2020retrieval}. A different decision arises once an answer
already exists and retrieval is used to reconsider it, as in answer-revision
systems~\citep{gao2023rarr,madaan2023selfrefine}: should the system return the
draft it has, or revise it in light of the evidence? Revision can repair an
incorrect answer, but it can also overwrite a correct one.

Draft confidence alone does not settle this choice. It estimates whether the
draft is correct. A correct draft can be preserved or harmed, and an incorrect
draft can be repaired or left wrong. Draft confidence indicates which of these
changes is possible, and predicting which one occurs requires modeling the
revision.

At the decision point studied here, the system has already produced a draft and
retrieved evidence but has not yet decoded a revision. The remaining question
is whether applying the refiner to this draft and this evidence is more likely
to repair the answer than to harm it. A policy that answers that question can
return some drafts without invoking the revision and reserve revision for cases
where it is predicted to help. The relevant signal may lie in the relationship
among the question, draft, evidence, and refiner.

For offline training and evaluation, we observe the outcome of both choices. A
generator first answers without retrieved context. A retriever then supplies
evidence to a fixed refiner, which produces a candidate revision from the same
question and draft. We grade both returning and revising for every example.
Figure~\ref{fig:outcome-matrix} separates the resulting \emph{preserved},
\emph{repair}, \emph{harm}, and \emph{unrecovered} cases.
Figure~\ref{fig:intro-examples} shows one example from each shaded cell under
the same retriever and refiner. We call the retriever, evidence prompt, and
refiner together a \emph{revision setup}, and the paired effect of that
revision setup its \emph{recoverability}.\footnote{The same term appears in
\citet{li2026raser} with a different decision point. See
Section~\ref{sec:related}.}

\begin{figure}[t]
\centering
\begin{tikzpicture}[x=1cm,y=1cm,every node/.style={inner sep=1.2pt},
    head/.style={font=\footnotesize},
    cell/.style={font=\fontsize{8}{9.5}\selectfont,anchor=center},
    outcome/.style={font=\fontsize{8}{9.5}\selectfont\bfseries,anchor=center}]
\path (-0.02,-0.02) rectangle (7.62,3.62);

\node[head,anchor=base] at (3.430,3.19) {Revision incorrect};
\node[head,anchor=base] at (6.200,3.19) {Revision correct};

\fill[RepairGreen!12] (4.815,1.870) rectangle (7.585,3.070);
\fill[HarmOrange!12] (2.045,0.670) rectangle (4.815,1.870);

\draw[black!30,line width=0.4pt] (2.045,0.670) rectangle (7.585,3.070);
\draw[black!30,line width=0.4pt] (4.815,0.670) -- (4.815,3.070);
\draw[black!30,line width=0.4pt] (2.045,1.870) -- (7.585,1.870);

\node[head,anchor=base east] at (1.945,2.420) {Draft incorrect};
\node[head,anchor=base east] at (1.945,1.220) {Draft correct};

\node[outcome,text=black!55] at (3.430,2.620) {UNRECOVERED};
\node[cell,text=black!55] at (3.430,2.320) {tie\quad$0$};
\node[outcome,text=RepairGreen] at (6.200,2.620) {REPAIR};
\node[cell,text=RepairGreen] at (6.200,2.320) {$+1$};

\node[outcome,text=HarmOrange] at (3.430,1.420) {HARM};
\node[cell,text=HarmOrange] at (3.430,1.120) {$-1$};
\node[outcome,text=black!55] at (6.200,1.420) {PRESERVED};
\node[cell,text=black!55] at (6.200,1.120) {tie\quad$0$};

\node[head,anchor=center] at (3.800,0.235)
  {Conditional recoverability $\Delta=\Pr(\mathrm{repair})-\Pr(\mathrm{harm})$};
\end{tikzpicture}
\caption{The revision decision is defined by paired outcomes. Evaluating both
the draft and its candidate revision places an example in one of four cells.
The diagonal cells are ties. Only the shaded off-diagonal cells distinguish
returning from revising. Draft correctness alone identifies only a row.}
\label{fig:outcome-matrix}
\end{figure}

\begin{figure}[t]
\centering
\footnotesize
\setlength{\tabcolsep}{0pt}
\renewcommand{\arraystretch}{1.05}
\begin{tabular}{@{}p{\columnwidth}@{}}
\toprule
\textcolor{RepairGreen}{\textbf{REPAIR}}\hfill{\fontsize{8}{9.5}\selectfont NQ-Open, DPR}\\
Who wrote the song \emph{Going to Kansas City}?\\
Charlie Christian $\rightarrow$ Jerry Leiber and Mike Stoller\\[1pt]
{\fontsize{8}{9.5}\selectfont\itshape\hspace{0.8em}[1] Kansas City (Leiber and Stoller song)
  \ldots ``Kansas City'' is a rhythm and blues song written by \textbf{Jerry
  Leiber and Mike Stoller} in 1952.}\\
\midrule
\textcolor{HarmOrange}{\textbf{HARM}}\hfill{\fontsize{8}{9.5}\selectfont TriviaQA, DPR}\\
Who was the second wife of Henry VIII?\\
Anne Boleyn $\rightarrow$ Anne of Cleves\\[1pt]
{\fontsize{8}{9.5}\selectfont\itshape\hspace{0.8em}[1] The Private Life of Henry VIII \ldots
  execution of his second wife, \textbf{Anne Boleyn} \ldots marries Jane Seymour
  \ldots He then weds a German princess, \textbf{Anne of Cleves}.}\\
\bottomrule
\end{tabular}
\caption{Revision repairs one draft and corrupts another. Two held-out
cases, each giving the question, draft $\rightarrow$ revision, and excerpts
of the top retrieved passage. In the repair case, the passage states the
correct answer and the refiner adopts it. In the harm case, the passage
confirms the draft, but the refiner switches to another name from the same
passage.}
\label{fig:intro-examples}
\end{figure}

We organize the study around four questions. First, how much repair and harm
does retrieval-conditioned revision create under different revision setups?
Second, does predicting the paired revision effect improve the return-or-revise
decision over predicting draft correctness? Third, what pre-revision
information and model capacity make that effect predictable? Finally, how does
the conclusion change when the policy may choose a draft-free standard-RAG
answer as well as the draft and its candidate revision? The first three
questions study the choice between two answers; the last tests whether the
conclusions hold when a third answer is available.

On 25,870 held-out NQ-Open~\citep{kwiatkowski2019natural,lee2019latent},
TriviaQA~\citep{joshi2017triviaqa}, and PopQA~\citep{mallen2023trust} examples,
revision repairs 9.32--13.42\% of drafts and corrupts 3.05--3.33\%. A policy
trained on the paired outcome improves on always revising by 1.10--1.33 points
(mean over three training seeds) and closes 35.9--41.4\% of the gap to an
oracle that sees both outcomes, while still applying 38--46\% of the harmful
revisions. In a matched comparison, the scorer trained on the revision effect
has greater area under the accuracy--revision-rate curve, which traces accuracy
as the policy revises a growing share of drafts, in all nine Llama setup--seed
fits, and at the selected thresholds it has higher mean accuracy in all three
setups, with significance across training runs for DPR. A diagnostic across
inputs and model classes finds the largest gain when a fine-tuned model reads
the complete question, draft, and evidence prompt.

However, the value of revision depends on which answers are available. A policy
choosing between the draft and a draft-free standard-RAG answer has higher mean
accuracy than one choosing between the draft and its candidate revision, and
adding candidate revision as a third option does not significantly raise
accuracy. For Llama, the standard-RAG answer is less accurate alone but more
useful as an alternative to the draft. This comparison limits the value of
candidate revision in the answer sets tested here: the paired revision effect
and the value of offering revision alongside other answers are different
quantities.

\paragraph{Contributions.}
This paper is an empirical study of this revision decision. Deciding on the
expected difference in correctness, rather than on first-stage confidence, is
established for model cascades and routing
\citep{jitkrittum2023cascade,ding2024hybrid,ong2025routellm,luo2026routelmt},
and paired outcomes with and without retrieval are used to decide when to
retrieve and which evidence to use
\citep{wang2023skr,tian2026retrievalutility,qu2025uplift}. We ask what this
target adds when the second answer is a revision of the first, conditioned on
the retrieved evidence. We make three contributions:
\begin{enumerate}
    \item Repair and harm rates are standard in self-correction studies
    \citep{huang2024cannot,kumar2025score}. We report them for
    retrieval-conditioned revision under three revision setups and one
    correctness judge, together with the headroom for choosing between
    returning and revising: the gap between the better fixed action and an
    oracle.
    \item We compare revision-effect prediction with draft-correctness
    prediction under matched training across three revision setups, showing a
    replicated advantage in area under the accuracy--revision-rate curve, and
    characterize how the available input and model class change policy
    performance.
    \item Choosing between a closed-book and a retrieval answer is an
    established form of adaptive retrieval \citep{mallen2023trust,wang2023skr}.
    We test that choice alongside revision: when a draft-free standard-RAG
    answer is also available, choosing between the draft and that answer
    (return-or-RAG) is stronger than return-or-revise, and adding candidate
    revision does not provide a significant third-action gain.
\end{enumerate}

\section{Paired Outcomes for the Revision Decision}
\label{sec:problem}

\paragraph{Limits of aggregate accuracy.}
Consider two revision setups that both raise accuracy by 3 points. The first
repairs 3\% of drafts and corrupts none. Nothing is at risk, and the right
policy is to revise everything. The second repairs 13\% and corrupts 10\%. The
average is identical, but in the second setup one draft in ten is correct
before the system runs and wrong after it, and there is clear value in a policy
that can recognize those cases. An aggregate compares two means and cannot tell
the two setups apart. We therefore observe, and then predict, the per-draft
outcome.

\paragraph{Paired outcome.}
For a question $q$, a generator produces an observed draft $a_d$. A fixed
revision setup $I$---a retriever, retrieval query, evidence prompt, and
refiner---produces a candidate revision $a_I$ from that same draft. A shared
semantic-equivalence judge grades both actions against the same accepted
answers, giving binary correctness labels $z_d$ and $z_I$. For information $x$
available before revision, we define the example-level outcome and conditional
recoverability as
\[
\begin{aligned}
r_I &= z_I-z_d\in\{-1,0,+1\},\\
\Delta_I(x) &= \mathrm{E}[r_I\mid x].
\end{aligned}
\]
Positive outcomes are repairs and negative outcomes are harms. Equivalently,
conditional recoverability is repair probability minus harm probability.

Unlike treatment-effect estimation from logged actions, and as in routing work
that scores both models on every training query
\citep{ding2024hybrid,luo2026routelmt}, this offline measurement observes both
outcomes for every example. The draft is already available, and the refiner can
be run once to construct the candidate revision; the same judge then grades
both answers against the same accepted answers. The paired label is measured
directly under that judge. Prediction is still required at the operational
decision point because the policy must act before revision decoding. This
design removes treatment-assignment ambiguity from evaluation, but the label
still depends on the chosen refiner, evidence, and correctness judge.

\paragraph{Decision rule and oracle gap.}
A policy revises when its predicted recoverability exceeds a threshold chosen
on development data. Draft confidence instead estimates $\Pr(z_d=1\mid x)$.
That identifies the row of Figure~\ref{fig:outcome-matrix} an example falls in.
The column, whether the revision repairs or harms that draft, requires the
paired outcome. \citet{jitkrittum2023cascade} draw the same distinction for
cascade deferral and find that confidence is often close to sufficient. Our
matched comparison measures how close it is for retrieval-conditioned revision.
An oracle that sees both labels takes the correct action whenever either
succeeds. In every setup we evaluate, always revising is the better fixed
action, so the remaining \emph{oracle gap} is exactly the harm rate. We report
a policy's accuracy gain over always revising and the fraction of this gap it
closes.

\paragraph{Revision setups.}
Because $I$ enters the definition, we always report recoverability for a stated
revision setup. Section~\ref{sec:setup} describes the three we evaluate.

\paragraph{Neighboring decisions.}
The decision point distinguishes this setting from several related uses of
outcome prediction. Selective prediction asks whether to trust the current
answer; adaptive RAG asks whether or how to retrieve; retrieval-utility
prediction asks whether context improves a fresh answer. Here we ask whether an
evidence-conditioned revision improves an observed draft, with the retrieved
evidence held fixed. Paired grading preserves the draft-specific transition
that aggregate accuracy and one-sided confidence discard.
Section~\ref{sec:related} places the setting in prior work.

\section{Predicting Revision Effect}
\label{sec:prediction}

\paragraph{Information inputs.}
We compare three inputs available before revision: the question alone; the
question and observed draft; and the complete revision prompt containing the
question, draft, instructions, and five retrieved passages. Every input ends
before revision decoding, and no decision input contains the candidate
revision, gold answer, judge verdict, or outcome label.

\paragraph{Model classes.}
Five model classes score return versus revise for every input. Four are
frozen-feature models: ridge regression on a frozen final state (Ridge); linear
and multilayer-perceptron classifiers on that state (Linear, MLP); and learned
attention pooling over all frozen token states (Attn-pool). The fifth is a
low-rank adapted (\emph{LoRA}) model~\citep{hu2022lora} that reads the
available input. Classifiers predict repair, harm, or tie (no change in
correctness) and use the softmax probability assigned to repair minus that
assigned to harm as their decision score. These probabilities are not
calibrated. A model paired with a threshold is a \emph{policy}: it selects an
action but does not generate the revision, and every policy chooses between the
same two answers. Appendix Table~\ref{tab:model-recipes} gives the exact
recipes.

\paragraph{Baselines.}
We compare against learned draft-correctness prediction and a \emph{Tian-style
baseline}, an engineered-feature regression adapted from retrieval-utility
prediction \citep{tian2026retrievalutility}. We also report a verbalized
$P(\mathrm{true})$ baseline~\citep{kadavath2022language}
(Appendix~\ref{sec:confidence-baselines}). Our Tian-style baseline keeps the
regression family and the utility-difference structure of the original, but
retargets the decision from retrieve-versus-not to revise-versus-return and
adapts the feature categories to our return-or-revise, pre-revision setting. We
do not reimplement \citeauthor{tian2026retrievalutility}'s learned
query-performance-prediction (QPP) and QualT5 document-quality components.
Verbalized $P(\mathrm{true})$ reads only the question and the draft, so it does
not depend on the revision setup; only its threshold is chosen per setup. The
draft-correctness target also does not depend on the setup, although that model
reads the retrieved passages.

\paragraph{Matched training targets.}
We also train two LoRA models that read the complete prompt: one predicts draft
correctness, the other the paired outcome. They share the backbone, adapter
settings, optimizer, training budget, and seeds. Output heads and checkpoint
selection follow their targets.

\section{Experimental Setup}
\label{sec:setup}

\paragraph{Revision setups.}
We evaluate Dense Passage Retrieval (DPR)~\citep{karpukhin2020dense},
BM25~\citep{robertson2009bm25,yang2017anserini}, and BM25 followed by MonoT5
reranking~\citep{nogueira2020monot5,bajaj2016msmarco}. Each returns five full
title--passage pairs from the same Wikipedia passage collection. The three
setups share the evidence prompt and refiner and differ only in retrieval.
These setups compare dense and sparse retrieval and the effect of reranking.

\paragraph{Data and evaluation.}
Training and development use 134,847 and 14,966 examples drawn from NQ-Open and
TriviaQA. Test contains 25,870 examples, of which 14,267 come from PopQA; PopQA
has no training split, so we use it only for evaluation
(Appendix~\ref{sec:reproducibility} gives every per-dataset split size).
Llama~3.1~8B Instruct~\citep{dubey2024llama} is the primary generator and
refiner. A deterministic Llama~3.3~70B semantic-equivalence judge
\citep{meta2024llama33,zheng2023judge} grades both actions. A second judge,
GPT-OSS-120B~\citep{openai2025gptoss}, rescores the same answers, and two
additional generator/refiner families, GPT-OSS-20B and OLMo~3~7B, test
sensitivity (Section~\ref{sec:robustness}). Each LoRA decision model is
fine-tuned from its family's generator/refiner (Llama~3.1~8B Instruct for the
main results). Models are fit on train. Checkpoints and thresholds are selected
on development and fixed before test. We train each learned approach with three
training seeds, holding the examples, drafts, candidate revisions, judge
labels, splits, prompts, and hyperparameters fixed, and report the mean $\pm$
sample standard deviation (SD) over the three runs. A dagger ($\dagger$) marks
a difference that passes the \emph{run-level test}: its mean over the three
runs is positive, and a two-sided paired $t$-test across the three run-matched
differences gives $p<0.05$. These tests are not corrected for multiple
comparisons. Deterministic baselines are fit once. The 45-cell input--model
grid and the per-dataset results in Appendix Table~\ref{tab:per-dataset} use a
single training run, without run-level tests. All intervals are paired 95\%
percentile bootstrap intervals over test examples, with 10,000 resamples drawn
within each dataset. They describe uncertainty over test examples, whereas the
SD describes variation across training seeds.

\paragraph{Offline supervision and online decisions.}
Candidate revisions are generated for every split so that paired labels can be
constructed and policies can be evaluated against the same answers. They are
supervision and evaluation artifacts, never policy inputs. At decision time,
the policy sees only the pre-revision inputs of Section~\ref{sec:prediction}
and either returns the observed draft or invokes the revision setup.

\section{Results}
\label{sec:results}

\begin{table*}[t]
\centering
\footnotesize
\setlength{\tabcolsep}{5pt}
\renewcommand{\arraystretch}{1.1}
\sbox0{\begin{tabular}{@{}lrrrrr@{\,$\pm$\,}lrr@{\,$\pm$\,}lr@{\,$\pm$\,}l@{}}
\toprule
\multicolumn{12}{@{}l}{\textbf{Panel A: Outcomes and decision performance}} \\
& \multicolumn{2}{c}{Outcome (\%)} & \multicolumn{5}{c}{Accuracy (\%)\,$\uparrow$} & \multicolumn{4}{c}{Policy vs.\ revise} \\
\cmidrule(lr){2-3}\cmidrule(lr){4-8}\cmidrule(lr){9-12}
& & & & & \multicolumn{2}{c}{} & & \multicolumn{2}{c}{Gain} & \multicolumn{2}{c}{Gap closed} \\
Revision setup & Repair\,$\uparrow$ & Harm\,$\downarrow$ & Return & Revise
  & \multicolumn{2}{c}{Policy} & Oracle & \multicolumn{2}{c}{(points)\,$\uparrow$}
  & \multicolumn{2}{c}{(\%)\,$\uparrow$} \\
\midrule
DPR & 9.32 & 3.05 & 47.38 & 53.65 & 54.91 & 0.18 & 56.70 & $+$1.26 & 0.18$^\dagger$ & 41.4 & 5.9 \\
BM25 & 10.58 & 3.33 & 47.38 & 54.63 & 55.96 & 0.22 & 57.96 & $+$1.33 & 0.22$^\dagger$ & 40.0 & 6.5 \\
BM25 $\rightarrow$ MonoT5 & 13.42 & 3.06 & 47.38 & 57.75 & 58.84 & 0.10 & 60.81 & $+$1.10 & 0.10$^\dagger$ & 35.9 & 3.3 \\
\end{tabular}}%
\usebox0\par\nointerlineskip
\begin{tabular*}{\wd0}{@{\extracolsep{\fill}}lr@{\extracolsep{0pt}\,$\pm$\,}l
  !{\extracolsep{\fill}}r@{\extracolsep{0pt}\,$\pm$\,}l
  !{\extracolsep{\fill}}r@{\extracolsep{0pt}\,$\pm$\,}l
  !{\extracolsep{\fill}}r@{\extracolsep{0pt}\,$\pm$\,}l@{}}
\midrule
\multicolumn{9}{@{}l}{\textbf{Panel B: Policy behavior}} \\
& \multicolumn{2}{c}{Revision rate} & \multicolumn{2}{c}{Repairs revised}
  & \multicolumn{2}{c}{Harms revised} & \multicolumn{2}{c}{Macro-averaged gain} \\
Revision setup & \multicolumn{2}{c}{(\% of examples)} & \multicolumn{2}{c}{(\%)\,$\uparrow$}
  & \multicolumn{2}{c}{(\%)\,$\downarrow$} & \multicolumn{2}{c}{(points)\,$\uparrow$} \\
\midrule
DPR & 38.8 & 5.1 & 96.0 & 1.1 & 46.4 & 8.9 & $+$1.16 & 0.16$^\dagger$ \\
BM25 & 59.5 & 13.3 & 94.6 & 0.4 & 43.0 & 5.3 & $+$1.34 & 0.21$^\dagger$ \\
BM25 $\rightarrow$ MonoT5 & 35.5 & 7.7 & 94.0 & 0.6 & 38.0 & 4.2 & $+$1.10 & 0.11$^\dagger$ \\
\bottomrule
\end{tabular*}\par
\caption{Measured recoverability and learned-policy results. Panel A: repair
and harm are the examples that revision changes from incorrect to correct and
from correct to incorrect. Accuracy is reported for always returning the draft,
always revising, the development-selected full-prompt LoRA policy, and the
return-or-revise oracle. Gain is the policy's accuracy minus always revising.
Gap closed divides this gain by the oracle gap. Panel B: the fractions of all
examples, of repairs, and of harms that the policy revises, and the
macro-averaged gain, which is the gain over always revising when NQ-Open,
TriviaQA, and PopQA are weighted equally rather than by example count. Policy
cells are the mean $\pm$ SD over three training seeds. $\dagger$ marks the
run-level test of Section~\ref{sec:setup}.}
\label{tab:outcomes}
\end{table*}

\subsection{Measured Repair and Harm}
\label{sec:repair-harm}

Table~\ref{tab:outcomes} shows that revision both repairs and harms drafts.
With DPR it repairs 9.32\% of drafts and corrupts 3.05\%. The other setups
repair up to 13.42\% and corrupt a similar share. The net effect is positive.
Always revising gains 6.27--10.37 accuracy points over returning. Harm,
however, is present in every setup, and the harm rate (3.05--3.33\%) is
therefore the oracle gap (Section~\ref{sec:problem}). Reranking yields more
repairs and a larger average revision effect, yet still corrupts about as many
drafts as DPR does. The harm case in Figure~\ref{fig:intro-examples} shows that
revision can overwrite a correct draft even when the top passage states the
draft's answer. The share of correct drafts harmed when no retrieved passage
contains a gold alias is highest after reranking, 21.1\% against 16.5\% for DPR
(Appendix Table~\ref{tab:harm-evidence-features}). Reranking increases repairs
but does not reduce harms. Most remaining examples are ties because the refiner
often returns the draft unchanged.

\subsection{The Learned Policy}
\label{sec:policy}

The full-prompt LoRA policy is fit once per revision setup and training seed,
with its checkpoint and threshold selected on development. In
Table~\ref{tab:outcomes}, it improves on always revising by 1.10--1.33 points
and closes 35.9--41.4\% of the oracle gap while revising 35.5--59.5\% of
examples. It revises 94.0--96.0\% of the repairs but also 38.0--46.4\% of the
harms, so harmful revision is the main remaining error.
Appendix~\ref{sec:harm-analysis} examines these cases and their retrieved
passages.

The accuracy gain is significant across seeds in every setup and spans only
0.98--1.53 points over the nine setup--seed fits, whereas the
development-selected revision rate is far less stable, spanning 29.9--74.4\%.
Accuracy can stay stable while the action rate varies because the accuracy
curves are relatively flat over a wide band of thresholds and because changing
the action on an unchanged answer does not change the returned text. For
example, the BM25 seed that revises 74\% of drafts changes the returned answer,
after normalization, on 29\% of its revisions, whereas the two seeds that
revise 49--55\% change it on 37--44\%, so the additional revisions fall almost
entirely on drafts the refiner returns unchanged.

Across the same three training runs, the policy also significantly outperforms
the Tian-style baseline in every setup (Appendix
Table~\ref{tab:tian-comparison}). That baseline revises 93.8--94.4\% of drafts
and scores within 0.1 points of always revising, so it is a weak comparison;
the matched draft-correctness method of Section~\ref{sec:training-target} is
the stronger one.

\subsection{Paired Outcome versus Draft Correctness}
\label{sec:training-target}

\begin{figure*}[t]
\centering
\includegraphics[width=\textwidth]{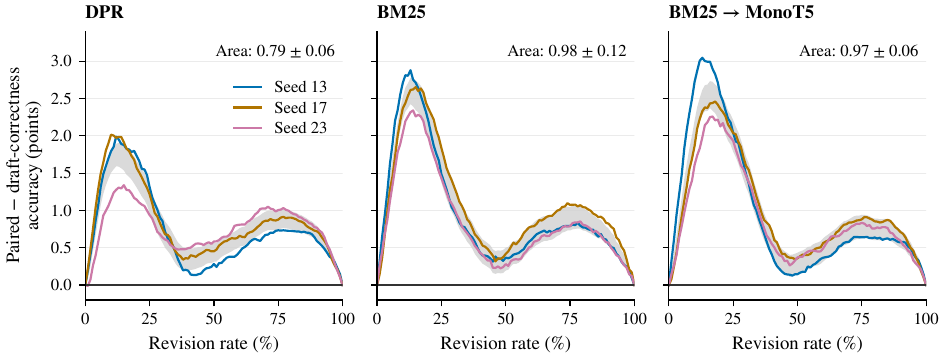}
\caption{Paired-outcome minus draft-correctness accuracy at matched revision
rates for the three revision setups. Lines are training seeds, and the gray
band is a 95\% bootstrap interval for the seed mean. Each area label gives the
mean $\pm$ SD of the integrated difference, in accuracy points.}
\label{fig:operating-curves}
\end{figure*}
\begin{table*}[t]
\centering
\footnotesize
\setlength{\tabcolsep}{5pt}
\renewcommand{\arraystretch}{1.1}
\begin{tabular}{@{}lr@{\,$\pm$\,}lr@{\,$\pm$\,}lr@{\,$\pm$\,}lr@{\,$\pm$\,}lr@{\,$\pm$\,}l@{}}
\toprule
& \multicolumn{4}{c}{Policy accuracy (\%)\,$\uparrow$} &
  \multicolumn{2}{c}{Paired $-$ draft\,$\uparrow$} &
  \multicolumn{4}{c}{Draft-correctness policy} \\
\cmidrule(lr){2-5}\cmidrule(lr){8-11}
Revision setup & \multicolumn{2}{c}{Draft-correctness} & \multicolumn{2}{c}{Paired-outcome} &
  \multicolumn{2}{c}{(points)} & \multicolumn{2}{c}{Revision rate (\%)} &
  \multicolumn{2}{c}{Harms revised (\%)\,$\downarrow$} \\
\midrule
DPR & 54.23 & 0.06 & 54.91 & 0.18 & $+$0.68 & 0.18$^\dagger$ & 57.8 & 2.0 & 74.1 & 3.6 \\
BM25 & 55.74 & 0.03 & 55.96 & 0.22 & $+$0.23 & 0.24 & 48.5 & 1.9 & 45.3 & 3.9 \\
BM25 $\rightarrow$ MonoT5 & 58.51 & 0.05 & 58.84 & 0.10 & $+$0.33 & 0.14 & 52.0 & 2.9 & 58.0 & 7.3 \\
\bottomrule
\end{tabular}
\caption{Matched paired-outcome and draft-correctness methods.
Cells are the mean $\pm$ SD over three training seeds. $\dagger$ marks the
run-level test of Section~\ref{sec:setup}. The paired-outcome column reuses
the
Table~\ref{tab:outcomes} policies.}
\label{tab:matched-targets}
\end{table*}

For Llama, Table~\ref{tab:matched-targets} and
Figure~\ref{fig:operating-curves} compare the two matched LoRA methods of
Section~\ref{sec:prediction}. The figure compares their rankings at equal
revision rates for all three training seeds.

The paired-outcome target has greater area under the accuracy--revision-rate
curve in all nine setup--seed fits, with maximum separation of 1.37--3.06
points, usually at low revision rates. The advantage holds at nearly every
revision rate; one DPR seed and one MonoT5 seed have small crossings near zero
revision rate, with deficits below 0.04 points.

At each policy's own development-selected threshold, the paired-outcome policy
exceeds the draft-correctness policy by 0.68 points on average for DPR, 0.23
for BM25, and 0.33 after MonoT5 reranking. The DPR difference is significant
across seeds; the BM25 difference is negative at one seed, and the MonoT5
difference is positive at every seed but narrowly misses the threshold
($p{=}0.058$). Both targets beat always revising. Of the 1.10--1.33-point gain
in Table~\ref{tab:outcomes}, draft correctness alone recovers 0.58--1.11 points
and the paired target adds the remaining 0.23--0.68, so the paired target
refines a decision that draft confidence can already make in part. The smaller
selected-threshold differences reflect the operating points chosen on
development, which lie in relatively flat regions of the curves.

Verbalized $P(\mathrm{true})$ confidence shows the same pattern more strongly.
It revises 85--93\% of drafts and gains only 0.03--0.09 points over always
revising (the DPR interval includes zero;
Appendix~\ref{sec:confidence-baselines}). Confidence ranks drafts by their own
correctness, which is a different question from whether revision will help
them.

A separate matched experiment changes only the output head and loss, with its
own three-way fits so that all four targets share one training and selection
protocol (Appendix~\ref{sec:target-formulations}). Two correctness heads, one
each for the draft and the revision, exceed the three-way repair/harm/tie head
by 0.13 points on average, and both have higher mean accuracy than a
scalar-utility target and a tie-aware preference target.

\input{figures/fig04_input_comparison}

\subsection{Inputs and Model Classes}

Figure~\ref{fig:input-comparison} compares every input within each model class
and revision setup in a single training run. The question alone sits near the
always-revise reference for every model class and revision setup, and carries
almost no usable signal. Adding the observed draft helps modestly. For each
revision setup, the best model class then closes between 6.8\% and 12.1\% of
the oracle gap. Large gains appear only when the fine-tuned LoRA policy reads
the complete question, draft, and evidence prompt, which closes 32.1--41.0\% in
this run (all 45 cells in Appendix Table~\ref{tab:input-model-grid};
Table~\ref{tab:outcomes} reports the three-seed mean of the same policy,
35.9--41.4\%). With the full prompt, the best frozen-feature model closes at
most 13.5\% of the gap, and the MLP does worse than with the question and draft
alone in every setup. Only the fine-tuned model gains substantially from the
full prompt. Because the model class changes together with the input, this
comparison does not separate what the policy reads from how it is trained. We
therefore retrain the full-prompt LoRA policy at the same three training seeds
with only its input changed (Appendix~\ref{sec:input-ablations}). When the
evidence is masked, or replaced by the evidence retrieved for a different
example, the gain over always revising falls from 1.10--1.33 to 0.18--0.25
points in every revision setup. Without the draft, it falls to 0.56--0.69
points.

\paragraph{Remaining errors.}
Row-level cases show why access to the complete prompt is useful but
insufficient. A retrieved passage can be topically close, fluent, and easy for
the refiner to follow while answering a different entity or time period. In the
examples in Figure~\ref{fig:intro-examples}, revision succeeds when the passage
supports the requested fact and fails when nearby evidence leads the refiner to
replace a correct draft with a plausible wrong answer. The aggregate lexical
associations in Appendix Table~\ref{tab:harm-evidence-features} support this
diagnosis. These results are consistent with a remaining problem that draft
confidence alone cannot resolve: whether the evidence answers the question
actually asked.

\subsection{Robustness}
\label{sec:robustness}

\begin{table*}[t]
\centering
\footnotesize
\setlength{\tabcolsep}{5pt}
\renewcommand{\arraystretch}{1.1}
\begin{tabular}{@{}llr@{\,$\pm$\,}lr@{\,$\pm$\,}l@{}}
\toprule
& & \multicolumn{2}{c}{Primary judge:} & \multicolumn{2}{c}{Second judge:} \\
Generator/refiner & Revision setup & \multicolumn{2}{c}{gain\,$\uparrow$} & \multicolumn{2}{c}{gain\,$\uparrow$} \\
\midrule
Llama~3.1~8B Instruct & DPR & $+$1.26 & 0.18$^\dagger$ & $+$1.37 & 0.19$^\dagger$ \\
& BM25 & $+$1.33 & 0.22$^\dagger$ & $+$1.34 & 0.21$^\dagger$ \\
& BM25 $\rightarrow$ MonoT5 & $+$1.10 & 0.10$^\dagger$ & $+$1.05 & 0.10$^\dagger$ \\
\addlinespace
GPT-OSS-20B & DPR & $+$0.12 & 0.03$^\dagger$ & $+$0.05 & 0.03 \\
& BM25 & $+$0.17 & 0.03$^\dagger$ & $+$0.06 & 0.03 \\
& BM25 $\rightarrow$ MonoT5 & $+$0.19 & 0.02$^\dagger$ & $+$0.12 & 0.04$^\dagger$ \\
\addlinespace
OLMo 3 7B & DPR & $+$2.74 & 0.11$^\dagger$ & $+$2.56 & 0.14$^\dagger$ \\
& BM25 & $+$2.07 & 0.02$^\dagger$ & $+$1.91 & 0.02$^\dagger$ \\
& BM25 $\rightarrow$ MonoT5 & $+$1.24 & 0.05$^\dagger$ & $+$1.17 & 0.04$^\dagger$ \\
\bottomrule
\end{tabular}\par
\caption{Policy gains across generator/refiner families and judges. Each entry
is the mean gain in points over always revising $\pm$ SD across three training
seeds. $\dagger$ marks the run-level test of Section~\ref{sec:setup}. Appendix
Table~\ref{tab:robustness-full} gives the oracle gaps and gap closed.}
\label{tab:robustness}
\end{table*}

PopQA is absent from train and development yet supplies 55.1\% of test
examples. In a single training run, the same policy and threshold improve
accuracy in all nine dataset--setup slices, with no per-dataset tuning. Because
PopQA dominates the pooled count, Table~\ref{tab:outcomes} also reports the
gain with the three datasets weighted equally: 1.10--1.34 points over three
seeds, against 1.10--1.33 pooled, so the pooled gain does not depend on the
evaluation-only dataset. In the single run, the weakest slice is DPR on
NQ-Open, at 14.6\% of the gap, even though NQ-Open supplies the largest share
of training data. It is also the DPR slice with the largest harm rate (4.18\%),
whereas evaluation-only PopQA closes 40.1\% under the same threshold. Appendix
Table~\ref{tab:per-dataset} gives the complete results for that run.

We also repeat the measurement and the policy comparison with the
GPT-OSS-20B~\citep{openai2025gptoss} and OLMo~3~7B~\citep{olmo2025olmo3}
generator/refiner families and a second automated judge. Our policy improves on
always revising in all nine generator/refiner and retrieval combinations, by
0.12--2.74 points under the primary judge and 0.05--2.56 points when the same
fixed decisions are rescored by the second judge. The GPT-OSS-20B gains are
roughly six to ten times smaller than Llama's, partly because its oracle gap is
smaller. Its refiner harms fewer drafts, so the oracle gap is 1.24--1.64 points
against 3.05--3.33 for Llama and 3.07--4.46 for OLMo (Appendix
Table~\ref{tab:robustness-full}). In relative terms it closes 9.9--11.6\% of
its gap, still below Llama's 35.9--41.4\%. For GPT-OSS-20B, all three
primary-judge gains reach run-level significance, but under the second judge
only the MonoT5-reranked setup does. Every Llama and OLMo gain is significant
under both judges. Table~\ref{tab:robustness} gives the comparison, and
Appendix Table~\ref{tab:robustness-full} gives the complete operating points.

As a check on the automated judge, one blind annotator who is not an author
independently labeled 480 draft--revision pairs and marked 14 of the 960
answers unsure. The annotator agrees with the primary judge on 93.2\% of drafts
($\kappa=0.86$) and 91.6\% of revisions ($\kappa=0.83$), and their three-way
outcome agreement (harm, tie, or repair) is 94.5\% ($\kappa=0.78$). Under the
human labels, revision repairs 8.5\% of the 470 pairs with two decisive labels
and harms 5.7\%. The pooled net effect is $+2.8$ points [$-0.6$, $+6.1$], and
the TriviaQA estimate is negative. On the same pairs, the primary judge gives
37 repairs and 26 harms (7.9\% and 5.5\%), so the human labels are slightly
more favorable to revision than the judge. Both labelings give a higher harm
rate than Table~\ref{tab:outcomes}, so the higher rate is a property of the
audit sample, on which the annotator and the judge agree. Appendix
Table~\ref{tab:human-diagnostic} gives the full results.

\subsection{Sensitivity to the Available Answers}
\label{sec:available-answers}

\begin{table*}[t]
\centering
\footnotesize
\setlength{\tabcolsep}{5pt}
\renewcommand{\arraystretch}{1.1}
\sbox0{\begin{tabular}{@{}lrrr@{\,$\pm$\,}lr@{\,$\pm$\,}lr@{\,$\pm$\,}lr@{\,$\pm$\,}l@{}}
\toprule
\multicolumn{11}{@{}l}{\textbf{Panel A: Fixed actions and learned policies: accuracy (\%)\,$\uparrow$}} \\
Revision setup & Revise & RAG & \multicolumn{2}{c}{Paired policy}
  & \multicolumn{2}{c}{\{Return,Revise\}} & \multicolumn{2}{c}{\{Return,RAG\}}
  & \multicolumn{2}{c}{\{Return,Revise,RAG\}} \\
\midrule
DPR & 53.65 & 49.12 & 54.91 & 0.18 & 55.03 & 0.18 & 57.08 & 0.10 & 56.93 & 0.25 \\
BM25 & 54.63 & 50.51 & 55.96 & 0.22 & 56.39 & 0.08 & 58.33 & 0.07 & 58.33 & 0.11 \\
BM25 $\rightarrow$ MonoT5 & 57.75 & 55.85 & 58.84 & 0.10 & 59.04 & 0.03 & 61.14 & 0.08 & 61.02 & 0.04 \\
\end{tabular}}%
\usebox0\par\nointerlineskip
\begin{tabular*}{\wd0}{@{\extracolsep{\fill}}lr@{\extracolsep{0pt}\,$\pm$\,}l
  !{\extracolsep{\fill}}r@{\extracolsep{0pt}\,$\pm$\,}l
  !{\extracolsep{\fill}}r@{\extracolsep{0pt}\,$\pm$\,}l@{}}
\midrule
\multicolumn{7}{@{}l}{\textbf{Panel B: Seed-matched accuracy contrasts (points)}} \\
& \multicolumn{2}{c}{\{Return,RAG\}} & \multicolumn{2}{c}{\{Return,RAG\}} & \multicolumn{2}{c}{\{Return,Revise,RAG\}} \\
Revision setup & \multicolumn{2}{c}{$-$ paired policy} & \multicolumn{2}{c}{$-$ \{Return,Revise\}}
  & \multicolumn{2}{c}{$-$ \{Return,RAG\}} \\
\midrule
DPR & $+$2.17 & 0.08$^\dagger$ & $+$2.05 & 0.09$^\dagger$ & $-$0.15 & 0.16 \\
BM25 & $+$2.37 & 0.16$^\dagger$ & $+$1.95 & 0.06$^\dagger$ & $+$0.00 & 0.08 \\
BM25 $\rightarrow$ MonoT5 & $+$2.30 & 0.04$^\dagger$ & $+$2.10 & 0.06$^\dagger$ & $-$0.12 & 0.11 \\
\bottomrule
\end{tabular*}\par
\caption{The available answers determine value. Test accuracy ($n{=}25{,}870$
per setup). Return is returning the draft, Revise the candidate revision, and
RAG a draft-free standard-RAG answer. Paired policy is the
Table~\ref{tab:outcomes} policy. Fixed actions are deterministic. Learned
cells are the mean $\pm$ SD over three training seeds. Panel B uses
seed-matched differences. $\dagger$ marks the run-level test of
Section~\ref{sec:setup}.}
\label{tab:action-sets}
\end{table*}

The results so far concern two answers: the returned draft and its candidate
revision. To test how much they depend on that pair, we train a LoRA model at
each of the same three training seeds to predict correctness for the returned
draft, a draft-free standard-RAG answer, and the candidate revision
(Table~\ref{tab:action-sets}). This per-action model reads the same
pre-revision prompt as the paired policy and fits one correctness head per
action; an action set restricts which heads compete. Standard RAG generates a
fresh answer from the question and retrieved evidence without conditioning on
the draft. It is a separate generation action, and the comparisons here do not
account for its generation cost. A policy choosing between return and standard
RAG exceeds the separately trained paired policy by 2.17--2.37 points on
average, although that comparison changes the prediction target as well as the
available answers. Holding each trained per-action model fixed, replacing
candidate revision with standard RAG still raises mean accuracy by 1.95--2.10
points across the three revision setups. Fixed standard RAG is weaker than
fixed revision in every setup but succeeds on more of the draft's failures, so
its value comes from failing on different examples. The contrast also holds
within each of the nine dataset--setup slices, where replacing candidate
revision with standard RAG gains 1.75--3.83 points and passes the run-level
test (Appendix Table~\ref{tab:action-sets-by-dataset}).

The same per-action model restricted to return-or-revise reaches 55.03, 56.39,
and 59.04\%, on average 0.12--0.43 points above the paired policy of
Table~\ref{tab:outcomes}. Section~\ref{sec:training-target} compares one
correctness head per action with the three-way target under matched training.

For Llama, adding candidate revision as a third option produces no significant
improvement. The mean differences are $-$0.15 points with DPR, $+$0.00 with
BM25, and $-$0.12 after reranking. Under the second judge, the DPR and reranked
losses have bootstrap intervals below zero (Appendix
Table~\ref{tab:second-judge}), but neither is significant across the three
training runs. No Llama dataset--setup slice moves by more than 0.24 points in
either direction. Candidate revision is the only correct answer on 181, 222,
and 323 of the 25,870 examples, raising the return-or-RAG oracle by 0.70, 0.86,
and 1.25 points. The learned three-action policies select only 7--24\% of those
unique revision wins across the nine fits. Table~\ref{tab:revision-switches}
separates successful switches to revision from switches that replace a correct
return-or-RAG choice with an incorrect revision. The lack of learned gain
reflects the balance of these switches.

\begin{table*}[t]
\centering
\footnotesize
\setlength{\tabcolsep}{5pt}
\renewcommand{\arraystretch}{1.1}
\begin{tabular}{@{}lrrrrr@{}}
\toprule
& Unique & Captured & Fixes to the & Damaging & Net change in \\
Revision setup & revision wins & unique wins & binary choice & switches & correct answers \\
\midrule
\multicolumn{6}{@{}l}{\textit{Llama 3.1 8B}} \\
DPR & 181 & 22.7 & 57.3 & 117.7 & $-$37.7 \\
BM25 & 222 & 34.3 & 44.0 & 77.7 & 0.7 \\
BM25 $\rightarrow$ MonoT5 & 323 & 53.0 & 77.0 & 161.7 & $-$31.7 \\
\addlinespace
\multicolumn{6}{@{}l}{\textit{OLMo 3 7B}} \\
DPR & 318 & 32.3 & 50.0 & 85.0 & $-$2.7 \\
BM25 & 360 & 38.7 & 71.0 & 162.3 & $-$52.7 \\
BM25 $\rightarrow$ MonoT5 & 479 & 82.7 & 69.3 & 192.7 & $-$40.7 \\
\bottomrule
\end{tabular}
\caption{What changes when revision becomes a third action. Unique wins count
examples where only revision is correct. Captured unique wins and fixes to the
binary choice are successful switches from the same per-action model's
return-or-RAG choice; fixes to the binary choice occur when an existing action
was correct but the binary selector chose the wrong one. Their sum minus
damaging switches equals the net change in correct answers. Switch counts are
means over three fits; unique wins are fixed. Each setup contains 25,870
examples.}
\label{tab:revision-switches}
\end{table*}

We run the standard-RAG comparison for Llama and OLMo; we did not generate
standard-RAG answers for GPT-OSS-20B. The same pattern holds, more sharply, for
the OLMo~3~7B family. For OLMo, we also generate and judge standard-RAG answers
in every setup and train the per-action model at the same three seeds. For
OLMo, fixed standard RAG beats fixed revision in every setup (41.57\% vs.\ 40.64\%
with DPR, 42.65\% vs.\ 41.77\% with BM25, 49.59\% vs.\ 47.39\% after
reranking), the reverse of Llama. Choosing between return and standard RAG
reaches $47.52 \pm 0.12$, $48.09 \pm 0.08$, and $53.41 \pm 0.02$\%, against
$43.54 \pm 0.07$, $43.94 \pm 0.06$, and $48.82 \pm 0.03$ for return and
revision from the same heads, a 4.0--4.6-point gap that is at least 3.4 points
in every dataset--setup slice. After reranking, the learned return-or-revise
policy trails simply always using standard RAG. Adding revision as a third
action changes mean accuracy by $-$0.01 $\pm$ 0.06, $-$0.20 $\pm$ 0.12, and
$-$0.16 $\pm$ 0.06 points over the three seeds. Paired recoverability measures
the value of a given revision setup. For OLMo, standard RAG uses the retrieved
evidence better than revision does.

\paragraph{Sensitivity of the action comparison.}
Our revision prompt tells the refiner to keep the draft unless the evidence
clearly supports a different answer. Replacing it with a neutral
keep-or-replace prompt lowers standalone Llama revision accuracy by 0.30--0.91
points but raises the return-or-revise oracle
(Appendix~\ref{sec:neutral-prompt}).

The policy contrasts also persist under a second judge: return-or-RAG leads
return-or-revise by 2.04--2.23 points for Llama, and paired prediction leads
draft-correctness prediction by 0.38--0.75 points. All six bootstrap intervals
for the three-seed means are positive (Appendix Table~\ref{tab:second-judge}).
Adding revision changes mean accuracy by $-0.12$ to $+0.01$ points.

\section{Related Work}
\label{sec:related}

\paragraph{Confidence and adaptive RAG.}
Selective prediction uses confidence to trade current-answer coverage for risk
\citep{chow1970optimum,geifman2017selective,kamath2020selective}. Adaptive RAG
decides whether, when, or how to retrieve using difficulty, uncertainty,
reflection tokens, or decoding state
\citep{mallen2023trust,wang2023skr,jiang2023active,asai2023self,jeong2024adaptive,
baek2025probing,moskvoretskii2025adaptive}. These methods act before or during
retrieval; our decision occurs after it. The draft and evidence are fixed, and
the choice is whether to apply the revision. We do not rerun Self-RAG or
Adaptive-RAG as baselines, because they change the generator or the retrieval
pipeline that our comparison holds fixed. The return-or-RAG action set in
Section~\ref{sec:available-answers} makes the same
answer-with-or-without-retrieval choice, but its model first reads the draft
and the retrieved passages. Rowen also acts after an initial answer, retrieving
to correct it when responses across languages or models disagree
\citep{ding2025rowen}. \citet{li2026raser} train RASER on the improvement from
escalating a one-shot RAG answer, and their multi-action variant predicts route
scores. They change the retrieval route; we hold the retrieved evidence fixed
and compare returning the observed draft with a revision conditioned on it.

\paragraph{Retrieval utility.}
Prior work predicts the gain from retrieval or the utility of retrieved
documents
\citep{dado2026rag,tian2026retrievalutility,dai2025seper,qu2025uplift,jiang2025gainrag}.
These methods include observed answer-quality differences with and without
context, as in \citet{tian2026retrievalutility,qu2025uplift}. Our two choices
begin from the same observed draft, and the evidence-conditioned answer is a
revision of that draft. Our Tian-style baseline (Section~\ref{sec:prediction})
adapts this utility-prediction structure to the pre-revision decision point.
Choosing among answerers is also studied through learning to defer, model
cascades, and LLM routing~\citep{madras2018predict,mozannar2020consistent,
chen2023frugalgpt,jitkrittum2023cascade,bouchard2026escalation,ding2024hybrid,ong2025routellm}.
RouteLLM, for example, learns from paired responses and win/tie/loss judgments,
including comparisons against gold answers. RouteLMT trains a router on the
gain of a larger translation model over a smaller one \citep{luo2026routelmt},
and concurrent work predicts rescue and harm separately to rank cascade
escalations \citep{wang2026srr}. Paired supervision before action selection is
shared with prior routing work. In these cascades the larger model answers the
input itself; here the second answer revises the first. Our comparison supplies
the observed draft and retrieved evidence to the policy and measures absolute
repair and harm under a fixed refiner.

\paragraph{Revision and conflicting evidence.}
Revision systems edit drafts from feedback
\citep{gao2023rarr,zhao2023verify,madaan2023selfrefine,wan2024adapt}.
\citet{huang2026counterrefine} similarly retrieve answer-conditioned support
and counterevidence in CounterRefine before a restricted keep-or-revise step.
\citet{wu2024clasheval} quantify conflict between internal priors and external
evidence in ClashEval, while situated-faithfulness and source-arbitration
methods decide how to use the two
\citep{huang2025situated,kim2025speak,zhu2026saber,wang2025astute}. Other work
documents harmful self-correction and cases where retrieved evidence overwrites
parametric knowledge
\citep{huang2024cannot,maekawa2024retrieval,li2024inconsistencies,ning2026revision,
liu2026selfcorrection,yoran2024robust}. Our study brings matched prediction
targets and fixed-scorer action comparisons to this revision setting. Like
utility and heterogeneous-effect prediction
\citep{qu2025uplift,athey2016recursive,kunzel2019metalearners}, it models an
outcome difference. Here both answers can be generated offline and graded,
making the difference observable under the judge. The empirical question is how
predictable that difference is and how the result changes when a standard-RAG
answer is an alternative to revision.

\section{Discussion}

\paragraph{What the paired target adds.}
Paired evaluation changes the prediction target and gives an explicit
return-or-revise ceiling. The selected policy recovers 35.9--41.4\% of it, and
the matched comparison shows that paired prediction refines draft-confidence
reasoning. A learned draft-correctness model already recovers part of the gain
over always revising, and predicting repair and harm adds 0.23--0.68 points on
average at the selected thresholds, with run-level significance for DPR
(Section~\ref{sec:training-target}). This small gain is consistent with the
finding of \citet{jitkrittum2023cascade} that confidence-based deferral often
suffices in model cascades. Draft correctness identifies which answers are at
risk, whereas the paired target also asks what the refiner will do to them. In
the separate target comparison, one correctness head per action does at least
as well as the three-class paired head, so predicting the revision's outcome is
useful under more than one formulation.

\paragraph{Where the useful signal comes from.}
In the input comparison, the question alone is weak, adding the draft helps
somewhat, and the largest gain comes from adapting the model on the complete
question, draft, and evidence prompt. With the model and training fixed, most
of that gain depends on the evidence retrieved for the question itself, and
part of it on the draft. This pattern is consistent with recoverability being a
relation among the question, draft, evidence, and refiner. In the matched
target comparison, the method trained for revision effect has higher mean
accuracy. Together, the two analyses indicate that the decision benefits from a
paired-outcome target and a model capable of interpreting the complete
pre-revision state.

\paragraph{Better retrieval does not remove the decision.}
More repair opportunities do not necessarily make individual revisions easier
to route. MonoT5 reranking produces the most repairs and the largest average
revision effect, yet the learned policy's gain over always revising is no
larger there, and harmful revisions remain. Better retrieval can enlarge the
opportunity without making the effect of a revision easier to predict.

\paragraph{The action set determines the value of revision.}
The defined recoverability $\Delta_I(x)$ compares the draft with its revision.
Adding another answer leaves that paired difference unchanged, but changes the
value of revision relative to the best available alternative. Candidate
revision is valuable when the only alternative is returning the draft. It adds
no significant accuracy gain, however, to a learned policy that can already
choose between the draft and a draft-free standard-RAG answer. For Llama,
standard RAG is weaker as a fixed action but fails on different examples, so
its complementarity matters more than its mean accuracy; for OLMo it is the
stronger fixed action and the gap between the two action sets widens to four
points. This result argues for evaluating the full action set at the decision
point: return, revise, standard RAG, and any other available action. Revision
does supply unique correct answers that raise the oracle, but learned gains
also depend on correcting return-or-RAG errors and avoiding damaging switches.

\paragraph{From accuracy to decision utility.}
Our development thresholds optimize final-answer accuracy. Under that
objective, a repair contributes one correct answer and a harm removes one, so
the two transitions enter symmetrically. A deployed system may instead assign
greater cost to corrupting a correct answer, charge for revision latency or
compute, or allow abstention and human escalation. Extending recoverability to
those settings requires specifying the utility of every available action,
selecting the operating point on development data, and testing whether that
utility transfers across revision setups and domains.

\section{Conclusion}

Deciding whether to revise requires estimating the effect of the revision.
Offline paired grading makes repair, harm, and the remaining oracle gap
observable under one correctness judge. A scorer trained on that paired outcome
has greater area under the accuracy--revision-rate curve than a matched
draft-correctness scorer in all nine Llama setup--seed fits, and the resulting
policy has higher mean return-or-revise accuracy in all three setups, with
run-level significance for DPR. The largest gain in the input diagnostic comes
from an adapted model that reads the full evidence prompt. The value of
revision is tied to the available answers. Once standard RAG is also available,
candidate revision provides no significant additional accuracy in the tested
policies. The decision to revise is worth learning only relative to the
available alternatives, and grading every available answer offline makes that
comparison measurable for training and evaluation.

\section*{Limitations}

Our study covers short-form open-domain question answering with three revision
setups and three generator/refiner families; we do not test long-form,
multi-hop, domain-specific, or high-stakes settings. In every family the same
checkpoint drafts and revises, and the models are open and no larger than
GPT-OSS-20B. Every setup retrieves five passages from one Wikipedia collection.
We do not test a refiner that differs from the drafter, larger models, other
corpora, deployed RAG pipelines, or a person revising the draft. PopQA is
evaluation-only and is 55\% of test; the per-dataset and macro-averaged results
reduce, but do not remove, this distribution-shift concern.

The input comparison and the matched target comparison both test complete
methods. The LoRA models differ from the frozen-feature models in adaptation,
parameter count, and optimization, and each matched target uses its own
prediction head and selection rule. The input ablations hold the LoRA model
fixed, but each ablation is trained separately, so they show which inputs the
policy needs to reach its accuracy rather than which inputs a trained policy
uses. Beyond the verbalized $P(\mathrm{true})$ baseline in
Appendix~\ref{sec:confidence-baselines}, we do not survey calibrated
confidence, semantic uncertainty, or other ways of modeling the two outcomes
\citep{guo2017calibration,kadavath2022language,farquhar2024semantic}. The
45-cell input--model grid, the per-dataset diagnostic, and the passage-title
and passage-order ablations use a single training run, and three training runs
give the run-level tests little power. Training seeds vary only policy training
(Section~\ref{sec:setup}), so they do not measure generation or judge-sampling
variance. The neutral prompt and second judge are each a single sensitivity
check, and both reuse the same test decisions without selecting new policies.

The refiner determines the oracle gap. Because the oracle gap equals the
refiner's harm rate in our setups (Section~\ref{sec:problem}), a refiner that
rarely overwrites a correct draft leaves a small gap for any policy to close,
as GPT-OSS-20B already shows (Section~\ref{sec:robustness}). For such a
refiner, paired measurement shows how little accuracy remains to gain, which is
worth knowing before training a policy.

Correctness labels come from automated semantic-equivalence judges. Rescoring
the same decisions with a second judge measures sensitivity to the judge, and
harm is the least stable of the paired outcomes. The accuracy gains keep their
sign under rescoring, although two GPT-OSS-20B gains lose run-level
significance, and the estimated fraction of the oracle gap closed changes with
the judge (Appendix Table~\ref{tab:robustness-full}). The human diagnostic uses
one blind annotator (Appendix~\ref{sec:human-diagnostic}), so it cannot
separate annotator idiosyncrasy from judge error or estimate inter-annotator
reliability, and it does not independently validate sub-point policy gains.

The standard-RAG comparison shows that candidate revision did not significantly
improve accuracy for the learned policies over the answer sets tested here, and
a nonsignificant gain is not an equivalence result. The shared per-action model
controls the action-set comparison, but its checkpoint-selection objective may
differ from that of a scorer optimized separately for each two-action set.

\section*{Ethical Considerations}

This is an accuracy study. The results do not establish deployment readiness.
Both failure modes, a missed repair and an applied harm, are consequential.
Deployment would require domain-specific correctness standards,
evidence-coverage checks, human escalation, and validation of every available
action. Wikipedia-derived evidence can be stale or unevenly distributed, so a
policy that is accurate on average may still deny repair systematically where
coverage is weak. All datasets, passages, and models used here are public.

\section*{Acknowledgments}
We thank Dr.\ Jim Davis for helpful feedback and discussion, and Lt Angie
Maravi Campos for data labeling. This work was supported in part by a grant of
computer time from the Department of Defense High Performance Computing
Modernization Program (HPCMP).

\bibliography{references}
\appendix
\setcounter{dbltopnumber}{3}
\renewcommand{\dbltopfraction}{0.90}
\renewcommand{\dblfloatpagefraction}{0.90}

\section{Supporting Analyses}

\subsection{Decision-Model Recipes}

Table~\ref{tab:model-recipes} lists the recipes. The main experiment fits the
first five model classes independently for each of three pre-revision inputs
and three revision setups in a single training run, 45 fits in total. Neural
optimizers use AdamW. Linear uses a $10^{-2}$ learning rate for 25 epochs, MLP
$10^{-3}$ for 10 epochs, and LoRA $10^{-4}$ for one epoch. The LoRA models
fine-tune Llama~3.1~8B Instruct, the generator itself; in the OLMo~3~7B and
GPT-OSS-20B runs they fine-tune that family's generator. The Tian-style
baseline is a separate complete-prompt comparison.

\begin{table*}[tp]
\centering
\scriptsize
\setlength{\tabcolsep}{4pt}
\renewcommand{\arraystretch}{1.1}
\begin{tabularx}{\textwidth}{@{}l>{\raggedright\arraybackslash\hsize=0.9\hsize}X
  >{\raggedright\arraybackslash\hsize=1.2\hsize}X
  >{\raggedright\arraybackslash\hsize=0.9\hsize}Xr@{}}
\toprule
Model class & Information read & Fit and score & Development selection & Trainable params. \\
\midrule
Ridge & Final frozen state after the available input &
  Standardized ridge regression on $y=z_I-z_d$; score is predicted change &
  Fixed convex fit; accuracy-maximizing threshold & 4,097 \\
Linear & Same final frozen state &
  $4096{\rightarrow}3$ head; unweighted CE; score is
  $P(\mathrm{repair})-P(\mathrm{harm})$ &
  Lowest $\Delta_I$ MSE checkpoint; accuracy-maximizing threshold & 12,291 \\
MLP & Same final frozen state &
  LayerNorm--$4096{\rightarrow}256$--GELU--dropout--$256{\rightarrow}3$;
  unweighted CE & Same & 1,057,795 \\
Attn-pool & All frozen final-layer token states in the available prefix &
  Learned single-query attention pooling and the same MLP head; unweighted CE &
  Same & 1,061,891 \\
LoRA model & Exact available pre-revision input, at most 4,096 tokens &
  Rank-16 LoRA on all attention and MLP projections plus a three-way head;
  unweighted CE & Lowest $\Delta_I$ MSE checkpoint; accuracy-maximizing threshold &
  $\approx$41.96M \\
\addlinespace
Matched draft correctness & Complete pre-revision prompt &
  Same LoRA recipe; two softmax logits; unweighted CE &
  Lowest draft Brier score; revise below accuracy-selected threshold &
  $\approx$41.96M \\
Per-action model & Complete prompt; Llama 4,096 / OLMo 8,192-token cap &
  Three correctness logits; independent sigmoid/BCE; eligible-action argmax &
  Highest three-action accuracy; earliest checkpoint on ties; no per-set refit &
  $\approx$41.96M \\
\addlinespace
Tian-style baseline & 21 complete-prompt retrieval, context, and draft features &
  OLS on $y$; score is predicted change &
  Fixed convex fit; accuracy-maximizing threshold & 22 \\
\bottomrule
\end{tabularx}
\caption{Decision-model recipes. Checkpoints and any policy thresholds are
selected on development. Threshold-selection ties favor fewer revisions;
three-action logit ties prefer return, standard RAG, then revision.}
\label{tab:model-recipes}
\end{table*}

\paragraph{Matched comparator and per-action model.}
The matched draft-correctness comparator uses two softmax logits and unweighted
cross-entropy. It selects the checkpoint minimizing development Brier score for
draft correctness, then selects a threshold maximizing development final-answer
accuracy. It revises when draft-correctness probability is strictly below that
threshold. The paired predictor instead selects its checkpoint by development
mean squared error (MSE) on the signed revision effect. The shared training
budget controls capacity and optimization, while these distinct selection
objectives remain part of the compared methods.

The per-action model uses independent correctness logits with unweighted binary
cross-entropy and sigmoid links, one each for return, standard RAG, and
revision. Checkpoint selection maximizes development accuracy of the
three-action argmax policy, retaining the earliest evaluated checkpoint on a
tie. Every restricted action set uses the same selected checkpoint. Actions are
chosen by the largest eligible logit; there is no set-specific threshold or
refit. Thus the two-action restrictions are controlled comparisons within that
model.

Every Llama LoRA model, including the comparator and the per-action model, uses
one epoch, effective batch 64 (four workers, two examples per device, eight
accumulation steps), fused AdamW, learning rate $10^{-4}$, weight decay 0.01,
cosine decay, warmup fraction 0.03, BF16, and gradient clipping at 1.0. LoRA
uses rank 16, scale 32, dropout 0.05, all seven attention/MLP projections, and
a trainable classification head; the GPT-OSS-20B models adapt the four
attention projections. We evaluate and save checkpoints every 250 updates.
Llama inputs are limited to 4,096 tokens and OLMo inputs to 8,192; overlength
inputs are rejected rather than truncated, and no test input exceeds its cap.

\subsection{Verbalized Confidence Baseline}
\label{sec:confidence-baselines}

The Llama~3.1~8B Instruct generator, with no adapter, is shown the question and
its own draft and asked to reply with one word, true or false, to whether the
draft is correct (Appendix~\ref{sec:prompt-templates} gives the prompt).
$P(\mathrm{true})$ is the probability of \emph{true} relative to \emph{false}
in that reply, from a single greedy generation. We convert $P(\mathrm{true})$
to a return-or-revise policy using only the 14,966-example development split,
then evaluate it once on all 25,870 test examples. It revises 85.3--93.1\% of
drafts and gains $+$0.027 points [$-$0.015, $+$0.070] over always revising with
DPR, $+$0.089 [$+$0.031, $+$0.147] with BM25, and $+$0.035 [$+$0.008, $+$0.066]
after MonoT5 reranking.

\subsection{Analysis of Harmful Revisions}
\label{sec:harm-analysis}
\begin{table*}[tp]
\centering
\footnotesize
\setlength{\tabcolsep}{5pt}
\renewcommand{\arraystretch}{1.1}
\begin{tabularx}{\textwidth}{@{}>{\raggedright\arraybackslash}Xrrr@{}}
\toprule
Retrieved-evidence feature & DPR & BM25 & BM25 $\rightarrow$ MonoT5 \\
\midrule
No passage carries a known gold alias: harm given correct draft
  & 439/2,666 (16.5) & 548/3,309 (16.6) & 421/1,992 (21.1) \\
$\geq 3$ passages carry a known gold alias: harm given correct draft
  & 149/5,532 (2.7) & 109/4,921 (2.2) & 162/6,775 (2.4) \\
Revision string occurs in evidence: share of harms
  & 676/790 (85.6) & 675/862 (78.3) & 681/792 (86.0) \\
Revision in evidence and no known gold alias: share of harms, pooled
  & \multicolumn{3}{c}{1,138/2,439 (46.7)} \\
\addlinespace
PopQA lexical-mismatch pattern: harm given correct draft
  & 221/256 (86.3) & 227/276 (82.2) & 223/255 (87.5) \\
PopQA lexical-mismatch pattern: coverage of harms
  & 221/411 (53.8) & 227/444 (51.1) & 223/422 (52.8) \\
PopQA lexical-mismatch pattern: repair given incorrect draft
  & 9/3,194 (0.3) & 17/2,770 (0.6) & 25/3,016 (0.8) \\
\bottomrule
\end{tabularx}
\caption{Retrieved-passage features associated with harmful revision.
Entries are count/denominator (percent). The pooled row counts harms whose
accepted answers have normalized aliases to check. It sums the three setups,
which share the same test questions, so we give no interval for it.}
\label{tab:harm-evidence-features}
\end{table*}

The two cases in Figure~\ref{fig:intro-examples} are both handled correctly.
The policy revises in the repair case and returns the draft in the harm case.

The row-level failures often look like faithful reading of the wrong passage.
One NQ-Open question asks when the Golden State Warriors won their first NBA
championship. The draft, 1947, is correct, but the retrieved titles concern the
2015--2018 teams and the refiner returns 2015. A PopQA question asks for the
screenwriter of \emph{The Terminal}. Passages about \emph{The Terminator} lead
the refiner from the correct Sacha Gervasi to William Wisher Jr. In both cases
the refiner follows the evidence it is given.
Table~\ref{tab:harm-evidence-features} gives the full-population rates behind
these patterns. Gold-alias matching and the check for the revision string in
the evidence are normalized lexical matches. The PopQA lexical-mismatch pattern
requires the revision in evidence, no passage title equal to the asked subject,
and no known gold alias in any passage. These test-set associations diagnose
the existing revision setups. They were not used to choose a method or
threshold. Apart from the PopQA lexical-mismatch pattern, the table reports the
harm side only; we do not give repair rates for the same evidence features.

\subsection{Retrieval-Utility Baseline}
\begin{table*}[tp]
\centering
\footnotesize
\setlength{\tabcolsep}{5pt}
\renewcommand{\arraystretch}{1.1}
\begin{tabular}{@{}lrrr@{\,$\pm$\,}l@{}}
\toprule
& & & \multicolumn{2}{c}{LoRA $-$ Tian-style} \\
Revision setup & Tian-style accuracy (\%) & Tian-style revision rate (\%) & \multicolumn{2}{c}{(points)\,$\uparrow$} \\
\midrule
DPR & 53.70 & 93.8 & $+$1.22 & 0.18$^\dagger$ \\
BM25 & 54.72 & 94.1 & $+$1.25 & 0.22$^\dagger$ \\
BM25 $\rightarrow$ MonoT5 & 57.80 & 94.4 & $+$1.04 & 0.10$^\dagger$ \\
\bottomrule
\end{tabular}\par
\caption{Tian-style baseline against the Table~\ref{tab:outcomes} policy.
The Tian-style baseline is deterministic. The difference is the mean $\pm$ SD
over three LoRA training seeds, and $\dagger$ marks the run-level test of
Section~\ref{sec:setup}.}
\label{tab:tian-comparison}
\end{table*}

Table~\ref{tab:tian-comparison} aligns the Tian-style baseline, our adaptation
of the retrieval-utility regression (Section~\ref{sec:prediction}), with the
Table~\ref{tab:outcomes} policy. The baseline's 22 coefficients stand against
about 42M trainable LoRA parameters. All thresholds maximize development
accuracy. The LoRA policy leads by 1.04--1.25 points in every setup. Because
the baseline's development threshold sends nearly every draft to revision
(Section~\ref{sec:policy}), this margin is close to the margin over always
revising in Table~\ref{tab:outcomes}.

\subsection{Additional Matched Training Targets}
\label{sec:target-formulations}
\begin{table*}[t]
\centering
\footnotesize
\setlength{\tabcolsep}{5pt}
\renewcommand{\arraystretch}{1.1}
\begin{tabular}{@{}lr@{\,$\pm$\,}lr@{\,$\pm$\,}lr@{\,$\pm$\,}l@{}}
\toprule
Target & \multicolumn{2}{c}{DPR} & \multicolumn{2}{c}{BM25} & \multicolumn{2}{c}{BM25 $\rightarrow$ MonoT5} \\
\midrule
Repair/harm/tie & 54.93 & 0.14 & 56.13 & 0.06 & 58.92 & 0.08 \\
Two correctness heads & 54.99 & 0.17 & 56.29 & 0.03 & 59.11 & 0.06 \\
Scalar utility regression & 54.74 & 0.13 & 55.77 & 0.14 & 58.46 & 0.13 \\
Tie-aware preference & 54.79 & 0.19 & 55.55 & 0.79 & 58.64 & 0.13 \\
\bottomrule
\end{tabular}
\caption{Separate matched target-formulation experiment. Accuracy (\%), mean
$\pm$ sample SD across three training seeds on the same 25,870 test questions.
All methods include the revision outcome in supervision. These fits are
separate from the Table~\ref{tab:outcomes} policies, so rows compare within
this table.}
\label{tab:target-formulations}
\end{table*}

Table~\ref{tab:target-formulations} reports a separate comparison with the
backbone, adapter, optimizer, epoch budget, and split protocol held fixed
within that experiment. The output head, link, and loss change with the target:
three-way repair/harm/tie classification; two binary correctness heads; scalar
regression on $z_I-z_d$; or a tie-aware preference target of $0$, $\tfrac12$,
and $1$. Every checkpoint and threshold was chosen on development. The
three-way target is refit inside this experiment so that all four targets share
one training and selection protocol; these fits differ from those in
Table~\ref{tab:outcomes}.

Two-head prediction exceeds three-way prediction by 0.13 points on average over
the nine setup--seed cells. Five of nine paired example-bootstrap intervals
exclude zero. Three-way and two-head prediction exceed scalar regression by
0.34 and 0.47 points, and the preference target by 0.33 and 0.47 points,
respectively. These mean contrasts describe nine cells that share the same test
data.

A correct estimate of the scalar difference $\mathrm{E}[z_I\mid
x]-\mathrm{E}[z_d\mid x]$ would suffice for the return-or-revise decision, so
this ordering reflects the fitted recipes; it does not show that scalar targets
lose information the decision needs.

\subsection{Complete Input and Model Comparison}
\begin{table*}[tp]
\centering
\footnotesize
\setlength{\tabcolsep}{5pt}
\renewcommand{\arraystretch}{1.1}
\begin{tabular}{@{}llrrr@{}}
\toprule
Revision setup & Model class & Question only & Question + draft & Question + draft + evidence \\
\midrule
DPR & Ridge & $-$1.0 & 2.4 & 7.7 \\
& Linear & $-$1.0 & $-$0.1 & 4.1 \\
& MLP & 0.3 & 6.8 & 2.4 \\
& Attn-pool & $-$0.5 & 2.7 & 2.7 \\
& LoRA & 0.1 & 5.8 & 35.8 \\
\addlinespace
BM25 & Ridge & $-$0.6 & 4.5 & 9.2 \\
& Linear & $-$1.3 & 7.0 & 13.5 \\
& MLP & 0.7 & 12.1 & 9.6 \\
& Attn-pool & $-$0.6 & $-$1.9 & 0.2 \\
& LoRA & $-$0.3 & 9.9 & 41.0 \\
\addlinespace
BM25 $\rightarrow$ MonoT5 & Ridge & $-$0.8 & 0.3 & 5.7 \\
& Linear & $-$0.6 & 1.9 & 4.5 \\
& MLP & 0.4 & 3.3 & 0.0 \\
& Attn-pool & $-$0.4 & 1.8 & 0.5 \\
& LoRA & $-$0.3 & 8.5 & 32.1 \\
\bottomrule
\end{tabular}\par
\caption{Every cell of Figure~\ref{fig:input-comparison}. Percentage of the
return-or-revise oracle gap closed over always revising for each revision
setup, model class, and pre-revision input (one training run).}
\label{tab:input-model-grid}
\end{table*}

Table~\ref{tab:input-model-grid} lists every cell of
Figure~\ref{fig:input-comparison}: the percentage of the oracle gap closed for
all 45 combinations of revision setup, model class, and pre-revision input on
the same 25,870 held-out examples. Each cell is trained independently on train
from the same training seed, with its checkpoint and threshold selected on
development. Figure~\ref{fig:input-comparison} gives the bootstrap intervals.

\subsection{Controlled Input Ablations}
\label{sec:input-ablations}
\begin{table*}[t]
\centering
\footnotesize
\setlength{\tabcolsep}{5pt}
\renewcommand{\arraystretch}{1.1}
\begin{tabular}{@{}lr@{\,$\pm$\,}lr@{\,$\pm$\,}lr@{\,$\pm$\,}l@{}}
\toprule
Input & \multicolumn{2}{c}{DPR} & \multicolumn{2}{c}{BM25} & \multicolumn{2}{c}{BM25 $\rightarrow$ MonoT5} \\
\midrule
Full prompt & 1.26 & 0.18 & 1.33 & 0.22 & 1.10 & 0.10 \\
Draft removed & 0.64 & 0.19$^\dagger$ & 0.69 & 0.49 & 0.56 & 0.17$^\dagger$ \\
Evidence shuffled & 0.18 & 0.04$^\dagger$ & 0.25 & 0.07$^\dagger$ & 0.23 & 0.05$^\dagger$ \\
Evidence masked & 0.18 & 0.02$^\dagger$ & 0.21 & 0.20$^\dagger$ & 0.23 & 0.01$^\dagger$ \\
\bottomrule
\end{tabular}
\caption{Gain over always revising, in accuracy points on all test examples,
when the full-prompt LoRA policy is retrained with part of its input removed or
replaced. Mean $\pm$ SD over three training seeds. $\dagger$ marks a drop from
the full prompt that passes the run-level test of Section~\ref{sec:setup}.}
\label{tab:input-ablations}
\end{table*}

Each ablation retrains the full-prompt LoRA policy of Table~\ref{tab:outcomes}
with the same recipe, training seeds, and development selection of checkpoint
and threshold, and changes only the prompt it reads, in train, development, and
test alike. \emph{Evidence masked} replaces every evidence token with one
neutral token, so the prompt keeps its length. \emph{Evidence shuffled} gives
each example the evidence retrieved for another example from the same dataset,
split, and revision setup, and no example keeps its own. \emph{Draft removed}
deletes the draft and keeps the rest of the prompt. The candidate revision and
the paired label are unchanged, so every policy chooses between the same two
answers. Table~\ref{tab:input-ablations} reports the gain over always revising.
In single runs, keeping only the five passage titles lowers accuracy by
0.41--0.89 points, and reversing or shuffling the passage order changes it by
at most 0.18 points. Two training seeds of the unchanged policy differ by up to
0.43 points.

\subsection{Results by Dataset}
\label{sec:per-dataset}
\begin{table*}[tp]
\centering
\footnotesize
\setlength{\tabcolsep}{5pt}
\renewcommand{\arraystretch}{1.1}
\begin{tabular}{@{}llrrrrrrr@{}}
\toprule
Revision & & Repair & Harm & & & Standard & & \\
setup & Dataset & (\%)\,$\uparrow$ & (\%)\,$\downarrow$ & Return\,$\uparrow$ & Revise\,$\uparrow$
  & RAG\,$\uparrow$ & Policy\,$\uparrow$ & Oracle\,$\uparrow$ \\
\midrule
DPR & NQ-Open & 15.04 & 4.18 & 47.40 & 58.25 & 56.65 & 58.86 & 62.44 \\
& TriviaQA & 7.32 & 2.85 & 78.22 & 82.68 & 77.24 & 83.89 & 85.54 \\
& PopQA & 8.99 & 2.88 & 30.10 & 36.22 & 31.46 & 37.37 & 39.10 \\
\addlinespace
BM25 & NQ-Open & 12.16 & 5.07 & 47.40 & 54.49 & 48.53 & 56.04 & 59.56 \\
& TriviaQA & 8.82 & 2.94 & 78.22 & 84.10 & 80.33 & 85.46 & 87.04 \\
& PopQA & 11.17 & 3.11 & 30.10 & 38.16 & 34.30 & 39.48 & 41.27 \\
\addlinespace
BM25 $\rightarrow$ MonoT5 & NQ-Open & 16.32 & 4.63 & 47.40 & 59.09 & 56.68 & 60.11 & 63.71 \\
& TriviaQA & 10.53 & 2.54 & 78.22 & 86.21 & 84.04 & 87.21 & 88.75 \\
& PopQA & 14.31 & 2.96 & 30.10 & 41.46 & 39.85 & 42.42 & 44.42 \\
\bottomrule
\end{tabular}\par
\caption{Results by dataset and revision setup (one training run). Repair and
harm rates, accuracy of the three fixed actions and the policy, and the
return-or-revise oracle, all in percent.}
\label{tab:per-dataset}
\end{table*}

Table~\ref{tab:per-dataset} gives results from one training run for every
dataset--setup slice under its development-selected policy and threshold
(Section~\ref{sec:robustness}). Gap closed is computed within each slice, as
defined in Section~\ref{sec:problem}. Always revising is the better fixed
action in all nine slices, exceeding both returning the draft and standard RAG.
Standard RAG is less accurate than revision in every slice, but can exceed
returning the draft. Its smallest deficit relative to revision is 1.60 points
on DPR/NQ-Open.

\subsection{Additional Robustness Checks}
\label{sec:robustness-checks}
\begin{table*}[tp]
\centering
\scriptsize
\setlength{\tabcolsep}{4pt}
\renewcommand{\arraystretch}{1.1}
\begin{tabular}{@{}llrr@{\,$\pm$\,}lr@{\,$\pm$\,}lrr@{\,$\pm$\,}lr@{\,$\pm$\,}l@{}}
\toprule
& & \multicolumn{5}{c}{Llama 3.3 70B judge} & \multicolumn{5}{c}{GPT-OSS-120B judge} \\
\cmidrule(lr){3-7}\cmidrule(lr){8-12}
Generator/refiner & Revision & Oracle gap & \multicolumn{2}{c}{Gain}
  & \multicolumn{2}{c}{Gap closed} & Oracle gap & \multicolumn{2}{c}{Gain}
  & \multicolumn{2}{c}{Gap closed} \\
& setup & (points)\,$\downarrow$ & \multicolumn{2}{c}{(points)\,$\uparrow$}
  & \multicolumn{2}{c}{(\%)\,$\uparrow$} & (points)\,$\downarrow$
  & \multicolumn{2}{c}{(points)\,$\uparrow$} & \multicolumn{2}{c}{(\%)\,$\uparrow$} \\
\midrule
Llama 3.1 8B Instruct & DPR & 3.05 & $+$1.26 & 0.18$^\dagger$ & 41.4 & 5.9 & 3.41 & $+$1.37 & 0.19$^\dagger$ & 40.2 & 5.6 \\
& BM25 & 3.33 & $+$1.33 & 0.22$^\dagger$ & 40.0 & 6.5 & 3.56 & $+$1.34 & 0.21$^\dagger$ & 37.6 & 6.0 \\
& BM25 $\rightarrow$ MonoT5 & 3.06 & $+$1.10 & 0.10$^\dagger$ & 35.9 & 3.3 & 3.44 & $+$1.05 & 0.10$^\dagger$ & 30.7 & 2.8 \\
\addlinespace
GPT-OSS-20B & DPR & 1.24 & $+$0.12 & 0.03$^\dagger$ & 9.9 & 2.3 & 1.67 & $+$0.05 & 0.03 & 2.9 & 1.6 \\
& BM25 & 1.51 & $+$0.17 & 0.03$^\dagger$ & 11.3 & 2.3 & 1.83 & $+$0.06 & 0.03 & 3.1 & 1.6 \\
& BM25 $\rightarrow$ MonoT5 & 1.64 & $+$0.19 & 0.02$^\dagger$ & 11.6 & 1.0 & 1.98 & $+$0.12 & 0.04$^\dagger$ & 5.9 & 1.9 \\
\addlinespace
OLMo 3 7B & DPR & 4.46 & $+$2.74 & 0.11$^\dagger$ & 61.5 & 2.4 & 4.43 & $+$2.56 & 0.14$^\dagger$ & 57.7 & 3.1 \\
& BM25 & 3.76 & $+$2.07 & 0.02$^\dagger$ & 55.1 & 0.5 & 3.75 & $+$1.91 & 0.02$^\dagger$ & 51.1 & 0.5 \\
& BM25 $\rightarrow$ MonoT5 & 3.07 & $+$1.24 & 0.05$^\dagger$ & 40.3 & 1.5 & 3.10 & $+$1.17 & 0.04$^\dagger$ & 37.6 & 1.2 \\
\bottomrule
\end{tabular}\par
\caption{Complete operating points behind Table~\ref{tab:robustness}.
Gain and gap closed are the mean $\pm$ SD over three training seeds.
$\dagger$ marks the run-level test of Section~\ref{sec:setup}.}
\label{tab:robustness-full}
\end{table*}

\begin{table*}[tp]
\centering
\footnotesize
\setlength{\tabcolsep}{5pt}
\renewcommand{\arraystretch}{1.1}
\begin{tabular}{@{}llrrrrrrrr@{}}
\toprule
& Primary & \multicolumn{2}{c}{\begin{tabular}[b]{@{}c@{}}Per-answer\\labels\end{tabular}} &
  \multicolumn{4}{c}{\begin{tabular}[b]{@{}c@{}}Second-judge label for each\\primary outcome ($n$)\end{tabular}} &
  & Retained \\
\cmidrule(lr){3-4}\cmidrule(lr){5-8}
Revision setup & outcome & Agree (\%) & $\kappa$ &
  Preserved & Repair & Harm & Unrecovered & Total & (\%) \\
\midrule
DPR & Preserved & 96.46 & 0.929 & 10{,}602 & 227 & 160 & 479 & 11{,}468 & 92.4 \\
& Repair & 96.59 & 0.932 & 20 & 2{,}257 & 5 & 129 & 2{,}411 & 93.6 \\
& Harm & & & 13 & 8 & 684 & 85 & 790 & 86.6 \\
& Unrecovered & & & 58 & 29 & 33 & 11{,}081 & 11{,}201 & 98.9 \\
\addlinespace
BM25 & Preserved & 96.46 & 0.929 & 10{,}562 & 238 & 133 & 463 & 11{,}396 & 92.7 \\
& Repair & 96.74 & 0.935 & 18 & 2{,}578 & 5 & 136 & 2{,}737 & 94.2 \\
& Harm & & & 17 & 4 & 747 & 94 & 862 & 86.7 \\
& Unrecovered & & & 56 & 29 & 37 & 10{,}753 & 10{,}875 & 98.9 \\
\addlinespace
BM25 $\rightarrow$ MonoT5 & Preserved & 96.46 & 0.929 & 10{,}590 & 276 & 169 & 431 & 11{,}466 & 92.4 \\
& Repair & 96.50 & 0.929 & 18 & 3{,}275 & 2 & 178 & 3{,}473 & 94.3 \\
& Harm & & & 21 & 3 & 679 & 89 & 792 & 85.7 \\
& Unrecovered & & & 57 & 45 & 39 & 9{,}998 & 10{,}139 & 98.6 \\
\bottomrule
\end{tabular}
\caption{Primary and second judge on the same test examples. The per-answer
label columns give binary correctness agreement and Cohen's $\kappa$ for the
draft (\emph{Preserved} line) and the revision (\emph{Repair} line). Drafts are
shared across setups.}
\label{tab:judge-agreement}
\end{table*}

\paragraph{Operating points.}
Table~\ref{tab:robustness-full} gives the oracle gap, gain, and gap closed
behind every entry of Table~\ref{tab:robustness}.

\paragraph{Judge agreement.}
Table~\ref{tab:judge-agreement} compares the primary and second judge on the
same test examples. Agreement on individual draft and revision verdicts
(per-answer agreement) is 96.46--96.74\% ($\kappa = 0.93$). Four-outcome
agreement, over preserved, repair, harm, and unrecovered, is 94.87--95.25\%
($\kappa = 0.92$). Harm is the least stable of the four outcomes. Only
85.7--86.7\% of primary-judge harms remain harms, against 98.6--98.9\% of
unrecovered cases, and most escaping harms become \emph{unrecovered}, meaning
the second judge disputes that the draft was ever correct. The second judge is
uniformly stricter, scoring 2.4--2.6 points fewer answers correct on every
branch, but it nevertheless finds \emph{more} harms overall (882, 922, and 889
against 790, 862, and 792), so the harms persist under the stricter judge.

\subsection{Single-Annotator Human Diagnostic}
\label{sec:human-diagnostic}
\begin{table*}[tp]
\centering
\footnotesize
\setlength{\tabcolsep}{5pt}
\renewcommand{\arraystretch}{1.1}
\sbox0{\begin{tabular}{@{}lrr@{\,/\,}lrr@{\,/\,}lrr@{\,/\,}l@{}}
\toprule
\multicolumn{10}{@{}l}{\textbf{Panel A: Single-annotator vs.\ judge agreement}} \\
& & \multicolumn{2}{c}{Draft agr.} & & \multicolumn{2}{c}{Rev.\ agr.} & & \multicolumn{2}{c}{Outcome agr.} \\
Comparison & $n_{\text{draft}}$ & \multicolumn{2}{c}{(\%) / $\kappa$} & $n_{\text{rev}}$
  & \multicolumn{2}{c}{(\%) / $\kappa$} & $n_{\text{pairs}}$
  & \multicolumn{2}{c}{(\%) / $\kappa$} \\
\midrule
Annotator vs.\ primary judge & 472 & 93.2 & 0.86 & 474 & 91.6 & 0.83 & 470 & 94.5 & 0.78 \\
Annotator vs.\ second judge & 471 & 93.8 & 0.88 & 471 & 91.7 & 0.83 & 467 & 95.1 & 0.80 \\
\end{tabular}}%
\usebox0\par\nointerlineskip
\begin{tabular*}{\wd0}{@{\extracolsep{\fill}}lrrrrrr@{\extracolsep{0pt}\ }l@{}}
\midrule
\multicolumn{8}{@{}l}{\textbf{Panel B: Human-labeled paired outcomes (decisive pairs)}} \\
& & Return & Revise & & & \multicolumn{2}{c}{$\Delta$ [95\% CI]} \\
Stratum & $n$ & acc.\ (\%) & acc.\ (\%) & Harm (\%) & Repair (\%) & \multicolumn{2}{c}{(points)} \\
\midrule
Pooled & 470 & 47.9 & 50.6 & 5.7 & 8.5 & $+$2.8 & [$-$0.6, $+$6.1] \\
NQ-Open & 157 & 34.4 & 42.0 & 3.8 & 11.5 & $+$7.6 & [$+$1.9, $+$13.8] \\
TriviaQA & 156 & 80.1 & 77.6 & 9.0 & 6.4 & $-$2.6 & [$-$8.9, $+$3.8] \\
PopQA & 157 & 29.3 & 32.5 & 4.5 & 7.6 & $+$3.2 & [$-$1.9, $+$8.4] \\
\bottomrule
\end{tabular*}\par
\caption{Single-annotator human diagnostic of paired outcomes. Outcome
agreement is computed on the three-way paired outcome (Harm, Tie, Repair).
$\Delta$ is Repair $-$ Harm in points with pair-level bootstrap 95\%
intervals.}
\label{tab:human-diagnostic}
\end{table*}

One annotator who is not an author, blind to whether each answer was a draft or
a revision, to the pairing, to the judges, and to metadata, labeled 960
candidates from 480 test pairs: 160 each from NQ-Open, TriviaQA, and PopQA. The
audit covers the DPR and BM25 setups. Each dataset--setup cell contributes 80
pairs, and no test example appears twice. Its estimates describe this
dataset-balanced audit sample, rather than the population-weighted test set.
Table~\ref{tab:human-diagnostic} reports per-answer agreement and three-way
outcome agreement, over harm, tie, and repair, between the annotator and each
judge. These comparisons leave out \emph{unsure} labels and invalid
second-judge verdicts, so $n$ varies; invalid primary-judge verdicts count as
incorrect, as everywhere else (Appendix~\ref{sec:reproducibility}). The
annotator marked 14 of the 960 answers unsure. Panel B covers pairs with two
decisive labels. Among the 470 decisive pairs, the primary judge scores 37
repairs and 26 harms. The annotator agrees with 29 of those repairs and 23 of
those harms, and labels 8 and 3 of them ties. The annotator also labels 11
repairs and 4 harms that the judge scores as ties. No pair moves between repair
and harm.

\subsection{Action Sets by Dataset}
\begin{table*}[t]
\centering
\scriptsize
\setlength{\tabcolsep}{4pt}
\renewcommand{\arraystretch}{1.1}
\begin{tabular}{@{}lrrr@{\,$\pm$\,}lr@{\,$\pm$\,}lr@{\,$\pm$\,}lr@{\,$\pm$\,}lr@{\,$\pm$\,}l@{}}
\toprule
Setup / dataset & Fixed $R$ & Fixed $G$ & \multicolumn{2}{c}{$D/R$} & \multicolumn{2}{c}{$D/G$}
  & \multicolumn{2}{c}{$D/G/R$} & \multicolumn{2}{c}{$D/G-D/R$} & \multicolumn{2}{c}{$D/G/R-D/G$} \\
\midrule
\multicolumn{13}{@{}l}{\textit{Llama 3.1 8B}; always return: 47.38\%} \\
DPR / pooled & 53.65 & 49.12 & 55.03 & 0.18 & 57.08 & 0.10 & 56.93 & 0.25 & 2.05 & 0.09 & $-$0.15 & 0.16 \\
\quad NQ-Open & & & 59.21 & 0.08 & 63.04 & 0.21 & 62.93 & 0.28 & 3.83 & 0.28$^\dagger$ & $-$0.11 & 0.11 \\
\quad TriviaQA & & & 84.31 & 0.05 & 86.10 & 0.11 & 86.08 & 0.05 & 1.79 & 0.16$^\dagger$ & $-$0.02 & 0.08 \\
\quad PopQA & & & 37.57 & 0.33 & 39.31 & 0.19 & 39.09 & 0.41 & 1.75 & 0.17$^\dagger$ & $-$0.23 & 0.24 \\
\addlinespace
BM25 / pooled & 54.63 & 50.51 & 56.39 & 0.08 & 58.33 & 0.07 & 58.33 & 0.11 & 1.95 & 0.06 & 0.00 & 0.08 \\
\quad NQ-Open & & & 56.35 & 0.16 & 58.51 & 0.38 & 58.55 & 0.24 & 2.16 & 0.36$^\dagger$ & 0.04 & 0.25 \\
\quad TriviaQA & & & 85.95 & 0.03 & 87.96 & 0.05 & 88.01 & 0.12 & 2.02 & 0.05$^\dagger$ & 0.04 & 0.08 \\
\quad PopQA & & & 39.83 & 0.10 & 41.68 & 0.05 & 41.66 & 0.09 & 1.85 & 0.12$^\dagger$ & $-$0.03 & 0.09 \\
\addlinespace
BM25 $\rightarrow$ MonoT5 / pooled & 57.75 & 55.85 & 59.04 & 0.03 & 61.14 & 0.08 & 61.02 & 0.04 & 2.10 & 0.06 & $-$0.12 & 0.11 \\
\quad NQ-Open & & & 60.35 & 0.03 & 62.82 & 0.29 & 62.95 & 0.36 & 2.47 & 0.27$^\dagger$ & 0.14 & 0.13 \\
\quad TriviaQA & & & 87.53 & 0.08 & 89.32 & 0.17 & 89.28 & 0.12 & 1.78 & 0.17$^\dagger$ & $-$0.04 & 0.05 \\
\quad PopQA & & & 42.75 & 0.04 & 44.93 & 0.09 & 44.69 & 0.05 & 2.18 & 0.05$^\dagger$ & $-$0.24 & 0.14 \\
\addlinespace
\multicolumn{13}{@{}l}{\textit{OLMo 3 7B}; always return: 29.17\%} \\
DPR / pooled & 40.64 & 41.57 & 43.54 & 0.07 & 47.52 & 0.12 & 47.51 & 0.09 & 3.98 & 0.13 & $-$0.01 & 0.06 \\
\quad NQ-Open & & & 46.05 & 0.15 & 52.50 & 0.17 & 52.62 & 0.18 & 6.45 & 0.31$^\dagger$ & 0.12 & 0.11 \\
\quad TriviaQA & & & 69.95 & 0.03 & 73.86 & 0.09 & 73.72 & 0.07 & 3.91 & 0.12$^\dagger$ & $-$0.13 & 0.11 \\
\quad PopQA & & & 28.11 & 0.11 & 31.50 & 0.16 & 31.52 & 0.15 & 3.39 & 0.09$^\dagger$ & 0.03 & 0.04 \\
\addlinespace
BM25 / pooled & 41.77 & 42.65 & 43.94 & 0.06 & 48.09 & 0.08 & 47.89 & 0.18 & 4.15 & 0.05 & $-$0.20 & 0.12 \\
\quad NQ-Open & & & 40.94 & 0.13 & 45.11 & 0.22 & 45.03 & 0.35 & 4.16 & 0.10$^\dagger$ & $-$0.07 & 0.12 \\
\quad TriviaQA & & & 71.81 & 0.06 & 76.17 & 0.14 & 75.98 & 0.08 & 4.35 & 0.10$^\dagger$ & $-$0.19 & 0.15 \\
\quad PopQA & & & 29.08 & 0.05 & 33.12 & 0.10 & 32.88 & 0.20 & 4.04 & 0.11$^\dagger$ & $-$0.25 & 0.10 \\
\addlinespace
BM25 $\rightarrow$ MonoT5 / pooled & 47.39 & 49.59 & 48.82 & 0.03 & 53.41 & 0.02 & 53.25 & 0.05 & 4.59 & 0.05 & $-$0.16 & 0.06 \\
\quad NQ-Open & & & 47.84 & 0.10 & 53.14 & 0.24 & 52.83 & 0.13 & 5.30 & 0.14$^\dagger$ & $-$0.31 & 0.11 \\
\quad TriviaQA & & & 76.54 & 0.07 & 81.03 & 0.04 & 80.95 & 0.03 & 4.48 & 0.03$^\dagger$ & $-$0.08 & 0.06 \\
\quad PopQA & & & 33.54 & 0.08 & 38.01 & 0.04 & 37.85 & 0.10 & 4.46 & 0.06$^\dagger$ & $-$0.16 & 0.07 \\
\bottomrule
\end{tabular}
\caption{Action-set comparison by family, revision setup, and dataset.
Accuracies are percentages; differences are points. $D$ returns the draft, $R$
revises it, and $G$ returns the draft-free standard-RAG answer. $D/R$, $D/G$,
and $D/G/R$ are learned policies; each learned cell is mean $\pm$ sample SD
over three training seeds. Within each family, setup, and seed, all three
action sets restrict the same per-action model; answers and labels are fixed.
Differences report the mean and SD of the within-seed difference, not a
difference of independent estimates. Each setup uses 3,610 NQ-Open, 7,993
TriviaQA, and 14,267 PopQA examples. No policy is retuned by dataset. $\dagger$
marks the run-level test of Section~\ref{sec:setup} for the dataset rows.}
\label{tab:action-sets-by-dataset}
\end{table*}

Table~\ref{tab:action-sets-by-dataset} compares the action sets for Llama and
OLMo in every revision setup, pooled and by dataset. In this and the following
action tables, $D$ returns the draft, $R$ applies the candidate revision, and
$G$ returns the draft-free standard-RAG answer; $D/R$ is the return-or-revise
action set, $D/G$ the return-or-RAG set, and $D/G/R$ the three-action set. The
dataset rows apply each per-action model to every dataset separately, with no
per-dataset fit, checkpoint, threshold, or action-set selection. Differences
are paired within model and example.

\subsection{Policy Contrasts Under a Second Judge}
\label{sec:second-judge}
\begin{table*}[t]
\centering
\footnotesize
\setlength{\tabcolsep}{5pt}
\renewcommand{\arraystretch}{1.1}
\begin{tabular}{@{}lr@{\,$\pm$\,}lr@{\,$\pm$\,}lr@{}}
\toprule
Revision setup & \multicolumn{2}{c}{Primary judge} & \multicolumn{2}{c}{Second judge} & Second-judge 95\% CI \\
\midrule
\multicolumn{6}{@{}l}{\textit{Paired outcome minus draft correctness}} \\
DPR & $+$0.68 & 0.18 & $+$0.75 & 0.20 & [$+$0.612, $+$0.880] \\
BM25 & $+$0.23 & 0.24 & $+$0.38 & 0.18 & [$+$0.215, $+$0.555] \\
BM25 $\rightarrow$ MonoT5 & $+$0.33 & 0.14 & $+$0.39 & 0.13 & [$+$0.240, $+$0.537] \\
\addlinespace
\multicolumn{6}{@{}l}{\textit{Return-or-RAG minus return-or-revise}} \\
DPR & $+$2.05 & 0.09 & $+$2.09 & 0.08 & [$+$1.823, $+$2.359] \\
BM25 & $+$1.95 & 0.06 & $+$2.04 & 0.01 & [$+$1.767, $+$2.312] \\
BM25 $\rightarrow$ MonoT5 & $+$2.10 & 0.06 & $+$2.23 & 0.09 & [$+$1.922, $+$2.533] \\
\addlinespace
\multicolumn{6}{@{}l}{\textit{Three actions minus return-or-RAG}} \\
DPR & $-$0.15 & 0.16 & $-$0.12 & 0.16 & [$-$0.198, $-$0.044] \\
BM25 & $+$0.00 & 0.08 & $+$0.01 & 0.12 & [$-$0.067, $+$0.081] \\
BM25 $\rightarrow$ MonoT5 & $-$0.12 & 0.11 & $-$0.10 & 0.10 & [$-$0.202, $-$0.004] \\
\addlinespace
\multicolumn{6}{@{}l}{\textit{Learned return-or-RAG policy minus return-or-revise oracle}} \\
DPR & $+$0.38 & 0.10 & $+$0.12 & 0.10 & [$-$0.187, $+$0.424] \\
BM25 & $+$0.37 & 0.07 & $+$0.17 & 0.07 & [$-$0.125, $+$0.482] \\
BM25 $\rightarrow$ MonoT5 & $+$0.33 & 0.08 & $+$0.06 & 0.06 & [$-$0.277, $+$0.394] \\
\bottomrule
\end{tabular}
\caption{Policy contrasts under both judges, in accuracy points. Mean $\pm$ SD
describes three policy-training seeds. The final column is a paired
example-bootstrap interval for the mean of those fixed fits. Every decision,
checkpoint, and threshold is unchanged. Both judges retain all 25,870 examples
per setup, counting invalid and ambiguous verdicts as incorrect. The matched
binary targets and the action-set comparisons are separate experiments.}
\label{tab:second-judge}
\end{table*}

We rescore the same test decisions with the second judge, GPT-OSS-120B at low
reasoning effort (Appendix~\ref{sec:artifacts}). The matched paired-target and
draft-correctness policies keep their development-selected thresholds, and all
action-set policies keep their argmax choices.

\subsection{Neutral Revision Prompt}
\label{sec:neutral-prompt}
\begin{table*}[t]
\centering
\footnotesize
\setlength{\tabcolsep}{5pt}
\renewcommand{\arraystretch}{1.1}
\begin{tabular}{@{}lrrrrr@{\ }l@{}}
\toprule
Revision setup & Neutral $R$ & Original oracle & Neutral oracle & Unique wins & \multicolumn{2}{c}{Oracle $-$ learned $D/G$} \\
\midrule
DPR & 52.74 & 56.70 & 57.22 & 198 & $+$0.14 & [$-$0.14, $+$0.42] \\
BM25 & 54.06 & 57.96 & 58.38 & 221 & $+$0.05 & [$-$0.23, $+$0.34] \\
BM25 $\rightarrow$ MonoT5 & 57.45 & 60.81 & 61.21 & 334 & $+$0.07 & [$-$0.24, $+$0.38] \\
\bottomrule
\end{tabular}
\caption{Prompt sensitivity on the same Llama test examples. Accuracy and
oracle cells are percentages. The last column is an accuracy difference in
points with a paired 95\% example-bootstrap interval for the mean of three
fixed policies. Neutral revision has lower standalone accuracy but a higher
oracle; the neutral oracle is above the learned seed mean in every setup. No
neutral-revision selector was fit.}
\label{tab:neutral-prompt}
\end{table*}

The primary revision prompt keeps the draft unless the evidence clearly
supports a different answer. The neutral prompt instead keeps the draft if it
is the answer best supported by the evidence and replaces it if a different
answer is better supported (Appendix~\ref{sec:prompt-templates} gives both
prompts). We generate neutral revisions with Llama and grade them with the
primary judge. No selector is retrained: the drafts, the standard-RAG answers,
and the learned return-or-RAG policy's decisions are unchanged.
Table~\ref{tab:neutral-prompt} compares neutral revision's standalone accuracy,
complementarity, and oracle with that learned return-or-RAG policy.

\section{Reproducibility Details}
\label{sec:reproducibility}

\paragraph{Splits.}
Train contains 79,029 NQ-Open and 55,818 TriviaQA examples. Development
contains 8,896 and 6,070. Test contains 3,610 NQ-Open, 7,993 TriviaQA, and
14,267 PopQA examples.

\paragraph{Training seeds.}
The three training seeds are 13, 17, and 23. Analyses that use a single
training run use the seed-13 fit: the 45-cell input--model grid, the and the
per-dataset results.

\paragraph{Retrieval and prompts.}
Appendix~\ref{sec:artifacts} gives the public resources, pinned model
revisions, determinism settings, and verbatim generation and judge prompts.

\paragraph{States.}
The final frozen state for the question alone is taken at the end of the exact
question-only input, before any draft or evidence token. For the question and
draft, it is taken in the candidate revision prompt at the last token before
the evidence. For the complete candidate revision prompt, which includes the
evidence and the generation prefix, it is taken at the last prompt token,
before the first revision token is decoded. All three states use Llama's final
RMS normalization.

\paragraph{Probability features.}
Answer probability is $\exp(\frac{1}{m}\sum_{t=1}^m\log p(a_t\mid a_{<t},x))$
over the exact parsed answer span. Context probability uses the analogous mean
over the exact evidence-token span in the prompt. Context features are computed
by teacher-forcing the exact tokenized prompts through the same Llama model in
vLLM. No answer is regenerated.

\paragraph{Invalid generations and judge parses.}
All 25,870 test examples remain in every reported accuracy calculation. Invalid
generations and judge parses are scored as explicit failures. Under the primary
judge, the drafts contain eight invalid and no ambiguous verdicts. The DPR,
BM25, and BM25 $\rightarrow$ MonoT5 revision branches contain, respectively,
six/one, five/zero, and three/one invalid/ambiguous verdicts. Seven drafts
lacking a parsed answer span across train/dev/test receive answer probability
zero in the Tian-style baseline's features and stay in the data.

\section{Artifacts, Prompts, and Determinism}
\label{sec:artifacts}
\begingroup\raggedright

\paragraph{Artifacts and availability.}
The upstream datasets, passage collection, and model families are public
research resources. No institution-internal data or closed commercial API model
is used. The experimental code, generated-answer corpus, trained policies, and
row-level evaluation outputs are not publicly released; the details below
describe the experiment for independent implementation. Questions and gold
aliases come from three public open-domain QA datasets: NQ-Open (Hugging Face
\texttt{nq\_open}), TriviaQA (\texttt{mandarjoshi/\allowbreak trivia\_qa},
configuration \texttt{rc.\allowbreak wikipedia.\allowbreak nocontext}), and
PopQA (\texttt{akariasai/\allowbreak PopQA}), which is evaluation-only and
contributes test rows exclusively. All three datasets, the passage collection,
and every model checkpoint are used as distributed by their original providers,
under their original licenses and stated terms of use, and are used here only
for the research purpose for which they were released. Splits are assigned
deterministically from the example identifier, with no private split file. The
identifier combines the dataset name with the source split and row index for
NQ-Open, the question ID for TriviaQA, and the ID field for PopQA. The source
datasets' own validation and test examples become our test split, and each
source training example is assigned to dev when the first eight hexadecimal
digits of the SHA-1 digest of its example identifier, read as an integer, are
congruent to $0$ modulo $10$, and to train otherwise. Both retrievers operate
over the public WikiDPR \texttt{psgs\_\allowbreak w100.\allowbreak multiset}
passage collection of 100-word Wikipedia passages with titles. DPR uses
\texttt{facebook/\allowbreak dpr-question\_encoder-\allowbreak multiset-base},
revision \texttt{5325e4ee\allowbreak 90643529\allowbreak 1d63046f\allowbreak
535476cb\allowbreak 3fc60d43}, through
Transformers~\citep{wolf2020transformers} \texttt{DPRQuestion\allowbreak
Encoder} and its fast tokenizer, with questions truncated at 256 tokens. It
searches the WikiDPR index \texttt{psgs\_w100.\allowbreak multiset.\allowbreak
HNSW128\_SQ8-IP-train.faiss}~\citep{johnson2021faiss}. The index corresponds to
WikiDPR revision \texttt{0ae24541\allowbreak 40a2d686\allowbreak
4475c83f\allowbreak 26e6dc9c\allowbreak d4ab9ce4}, and every split uses this
encoder snapshot and index file. BM25 uses Anserini's \texttt{Search\allowbreak
Collection}~\citep{yang2017anserini} over a Lucene index built from the same
title-plus-text passages, invoked with the \texttt{-bm25} ranker at its default
parameterization. Tokenization, lowercasing, and stopword handling are
inherited from that Lucene index's analyzer and were not modified. The reranked
setup retrieves 100 BM25 candidates and retains the top five under
\texttt{castorini/\allowbreak monot5-base-\allowbreak msmarco-10k}, revision
\texttt{f15657ab\allowbreak 3d2a5dd0\allowbreak b9a30c8c\allowbreak
0b6a0a73\allowbreak c9cb5884}. Generation uses public model checkpoints pinned
by commit revision: \texttt{meta-llama/\allowbreak Llama-3.1-\allowbreak
8B-Instruct} (revision \texttt{0e9e39f2\allowbreak 49a16976\allowbreak
918f6564\allowbreak b8830bc8\allowbreak 94c89659}) as both generator and
refiner; \texttt{openai/\allowbreak gpt-oss-20b} (revision
\texttt{6cee5e81\allowbreak ee839178\allowbreak 06bbde32\allowbreak
0786a8fb\allowbreak 61efebee}) and \texttt{allenai/\allowbreak
Olmo-3-7B-Instruct} (revision \texttt{6e5971d9\allowbreak eba42665\allowbreak
f5bd5a0f\allowbreak cf047f29\allowbreak 9ce1dccc}) as additional
generator/refiner families; \texttt{meta-llama/\allowbreak
Llama-3.3-\allowbreak 70B-Instruct} (revision \texttt{6f6073b4\allowbreak
23013f6a\allowbreak 7d4d9f39\allowbreak 144961bf\allowbreak bfbc386b}) as the
primary semantic-equivalence judge; \texttt{openai/\allowbreak gpt-oss-120b}
(revision \texttt{b5c939de\allowbreak 8f754692\allowbreak c1647ca7\allowbreak
9fbf85e8\allowbreak c1e70f8a}, reasoning effort \texttt{low}) as the second
judge. All models are served with vLLM 0.25.1~\citep{kwon2023vllm} on top of
Transformers 5.5.4 and PyTorch 2.11.0. The frozen-feature models are fit with
scikit-learn 1.8.0, NumPy 2.3.5, and SciPy 1.17.1. The 70B and 120B judges run
with tensor parallelism over eight GPUs. The 8B generator runs on a single GPU.

\paragraph{Compute budget.}
All GPU experiments ran on NVIDIA H200 GPUs of a Department of Defense HPC
Modernization Program cluster. Reproducing the reported experiments once is
estimated to require approximately 900 H200 GPU-hours. This estimate counts
each generation, judging, and policy-training stage once, with all three
training seeds sharing the same generations and verdicts.

\paragraph{Determinism.}
Answer generation for Llama and OLMo uses a single greedy sample with
\texttt{temperature=0.0}, \texttt{top\_p=1.0}, \texttt{top\_k=0}, and
\texttt{seed=13}. GPT-OSS answer generation uses Harmony, the GPT-OSS chat
format, and samples with seed 13 at \texttt{temperature=1.0} and
\texttt{top\_p=1.0}, the sampling settings OpenAI recommends for
GPT-OSS.\footnote{\url{https://github.com/openai/gpt-oss}} The primary Llama
judge is greedy. The GPT-OSS second judge uses the same Harmony sampling
settings. Both judges are served locally, and no external or commercial API is
called.

\subsection{Prompt Templates}
\label{sec:prompt-templates}

Each block below is the exact user-message content produced by our generation
code, with substituted fields shown in braces. The strings are Python format
templates: \texttt{\{question\}}, \texttt{\{candidate\_answer\}},
\texttt{\{evidence\}}, \texttt{\{gold\_aliases\}}, and \texttt{\{draft\}} are
the only substituted fields; the braces in the judge prompt's JSON example are
literal. On the non-Harmony Llama and OLMo paths, the string is placed as a
single \texttt{user} turn with no system message and wrapped by the served
model's own chat template with a generation prompt appended. GPT-OSS uses the
same displayed user content inside its Harmony conversation. Our code adds
Harmony's system message to set reasoning effort. In the blocks, a flush-left
line is a newline in the template and an indented continuation is only column
wrapping.

\paragraph{Draft prompt.}
\begin{promptblock}
\pline{Answer the question using only the question and your existing knowledge.}
\pline{\strut}
\pline{Return a short answer string, not a sentence. Use the minimal unambiguous answer that directly satisfies the question.}
\pline{For people, places, organizations, works, dates, numbers, and titles, return only the name, date, number, or title.}
\pline{If the question asks for multiple answers, separate them with semicolons.}
\pline{Do not explain, cite sources, hedge, or output "unknown". If unsure, give your best concise answer.}
\pline{\strut}
\pline{Return exactly one line in this format:}
\pline{Final answer: <answer>}
\pline{\strut}
\pline{Question: \{question\}}
\end{promptblock}

\paragraph{Standard RAG prompt.}
This branch conditions on the question and evidence but not on the draft.
\begin{promptblock}
\pline{Answer the question using the provided evidence.}
\pline{\strut}
\pline{Return a short answer string, not a sentence. Use the minimal unambiguous answer that directly satisfies the question.}
\pline{Use the evidence as authoritative when it clearly answers the question.}
\pline{If the evidence contains several aliases or descriptions for the same answer, return the most canonical short form.}
\pline{If the evidence is incomplete or irrelevant, still return your best concise answer; do not output "unknown".}
\pline{\strut}
\pline{Do not explain, cite sources, quote evidence, or give multiple alternatives.}
\pline{\strut}
\pline{Return exactly one line in this format:}
\pline{Final answer: <answer>}
\pline{\strut}
\pline{Question: \{question\}}
\pline{\strut}
\pline{Evidence:}
\pline{\{evidence\}}
\end{promptblock}

\paragraph{Candidate revision prompt.}
\begin{promptblock}
\pline{You are revising a candidate answer to an open-domain question.}
\pline{\strut}
\pline{Use the retrieved evidence only to decide whether to KEEP or REPLACE the candidate answer.}
\pline{\strut}
\pline{Rules:}
\pline{- KEEP the candidate answer exactly as written if the evidence supports it.}
\pline{- KEEP the candidate answer exactly as written if the evidence is insufficient, irrelevant, ambiguous, conflicting, or does not answer the question.}
\pline{- REPLACE the candidate answer only when the evidence clearly supports a different answer.}
\pline{- When replacing, return only the minimal unambiguous short answer supported by the evidence.}
\pline{- Do not abstain, refuse, discuss the evidence, cite sources, hedge, output "unknown", or provide multiple alternatives.}
\pline{- Return a short answer string, not a sentence.}
\pline{\strut}
\pline{Return exactly one line in this format:}
\pline{Final answer: <answer>}
\pline{\strut}
\pline{Question: \{question\}}
\pline{\strut}
\pline{Candidate answer: \{candidate\_answer\}}
\pline{\strut}
\pline{Evidence:}
\pline{\{evidence\}}
\end{promptblock}

\paragraph{Neutral revision prompt.}
Used only for the prompt comparison in Appendix~\ref{sec:neutral-prompt}. It
differs from the candidate revision prompt only in its KEEP and REPLACE rules.
\begin{promptblock}
\pline{You are revising a candidate answer to an open-domain question.}
\pline{\strut}
\pline{Use the retrieved evidence only to decide whether to KEEP or REPLACE the candidate answer.}
\pline{\strut}
\pline{Rules:}
\pline{- KEEP the candidate answer exactly as written if it is the answer best supported by the evidence.}
\pline{- REPLACE the candidate answer if a different answer is better supported by the evidence.}
\pline{- When replacing, return only the minimal unambiguous short answer supported by the evidence.}
\pline{- Do not abstain, refuse, discuss the evidence, cite sources, hedge, output "unknown", or provide multiple alternatives.}
\pline{- Return a short answer string, not a sentence.}
\pline{\strut}
\pline{Return exactly one line in this format:}
\pline{Final answer: <answer>}
\pline{\strut}
\pline{Question: \{question\}}
\pline{\strut}
\pline{Candidate answer: \{candidate\_answer\}}
\pline{\strut}
\pline{Evidence:}
\pline{\{evidence\}}
\end{promptblock}

\paragraph{Evidence block.}
The \texttt{\{evidence\}} field of the standard-RAG and candidate-revision
prompts is assembled from the retrieved passages before formatting. Passages
are numbered from one in rank order. Each passage contributes its title and
full text, right-stripped, and consecutive passages are separated by a blank
line. A passage with an empty title contributes only the bracketed index and
its text. We pack the complete top-five title-plus-text passages, with no
character cap.
\begin{promptblock}
\pline{[1] <title of passage 1>}
\pline{<full text of passage 1>}
\pline{\strut}
\pline{[2] <title of passage 2>}
\pline{<full text of passage 2>}
\end{promptblock}

\paragraph{Semantic-equivalence judge prompt.}
Both judges receive the identical prompt. \texttt{\{gold\_aliases\}} and
\texttt{\{candidate\_answer\}} are inserted as JSON values, so a candidate
answer appears quoted and the alias set appears as a JSON array.
\begin{promptblock}
\pline{You are judging short-answer QA correctness.}
\pline{\strut}
\pline{Given a question, accepted gold aliases, and a candidate answer, decide whether}
\pline{the candidate is semantically equivalent to one accepted answer.}
\pline{\strut}
\pline{Allow aliases, abbreviations, date formats, equivalent named entities, minor}
\pline{formatting differences, and answer-bearing phrases.}
\pline{\strut}
\pline{Do not give credit for merely related entities, partial overlaps, broader or}
\pline{narrower answers, or unsupported elaborations.}
\pline{\strut}
\pline{Return one JSON object and no other text. Use exactly these keys:}
\pline{\{"verdict":"equivalent"\textbar{}\allowbreak"not\_equivalent"\textbar{}\allowbreak"ambiguous"\textbar{}\allowbreak"invalid",\allowbreak"confidence":0.0,\allowbreak"matched\_alias":null,\allowbreak"notes":""\}}
\pline{\strut}
\pline{Question: \{question\}}
\pline{Gold aliases: \{gold\_aliases\}}
\pline{Candidate answer: \{candidate\_answer\}}
\end{promptblock}

\paragraph{$P(\mathrm{true})$ prompt.}
Used only for the confidence baseline in
Appendix~\ref{sec:confidence-baselines}. \texttt{\{draft\}} is the Llama draft
answer.
\begin{promptblock}
\pline{You are checking whether a proposed answer to a question is correct.}
\pline{\strut}
\pline{Question: \{question\}}
\pline{Proposed answer: \{draft\}}
\pline{\strut}
\pline{Is the proposed answer correct?}
\pline{Reply with exactly one word, either true or false. Do not explain.}
\end{promptblock}

\paragraph{Validity handling.}
The locally served judge uses grammar-constrained JSON with one verdict in
\{\texttt{equivalent}, \texttt{not\_equivalent}, \texttt{ambiguous},
\texttt{invalid}\}. A deterministic parser extracts the required one-line
\texttt{Final answer:} field from every generation. Malformed, truncated,
ambiguous, and invalid outputs are scored incorrect.
\par\endgroup

\end{document}